\documentclass[preprints,article,accept,moreauthors,pdftex]{Definitions/mdpi} 
\firstpage{1} 
\pubvolume{1}
\issuenum{1}
\articlenumber{0}
\pubyear{2021}
\copyrightyear{2021}
\externaleditor{Academic Editor: Hector Zenil}
\datereceived{19 March 2021} 
\dateaccepted{26 April 2021} 
\datepublished{} 
\hreflink{https://doi.org/}

\Title{Forward and Backward Bellman Equations Improve the Efficiency of the EM Algorithm for DEC-POMDP}
\TitleCitation{Forward and Backward Bellman Equations Improve the Efficiency of the EM Algorithm for DEC-POMDP}

\Author{Takehiro Tottori 
 $^{1,}$* and Tetsuya J. Kobayashi $^{1,2,3,4}$}

\AuthorNames{Takehiro Tottori, Tetsuya J. Kobayashi}

\AuthorCitation{Tottori, T.; Kobayashi, T.J.}

\address{%
$^{1}$ \quad Department of Mathematical Informatics, Graduate School of Information Science and Technology,\linebreak The University of Tokyo, Tokyo 113-8654, Japan\\ 
$^{2}$ \quad Department of Electrical Engineering and Information Systems, Graduate School of Engineering,\linebreak The University of Tokyo, Tokyo 113-8654, Japan \\
$^{3}$ \quad Institute of Industrial Science, The University of Tokyo, Tokyo 153-8505, Japan\\
$^{4}$ \quad Universal Biology Institute, The University of Tokyo, Tokyo 113-8654, Japan
}

\corres{Correspondence: takehiro\_tottori@sat.t.u-tokyo.ac.jp} 

\abstract{Decentralized partially observable Markov decision process (DEC-POMDP) models sequential decision making problems by a team of agents. 
Since the planning of DEC-POMDP can be interpreted as the maximum likelihood estimation for the latent variable model, 
DEC-POMDP can be solved by the EM algorithm. 
However, in EM for DEC-POMDP, the forward--backward algorithm needs to be calculated up to the infinite horizon, 
which impairs the computational efficiency. 
In this paper, we propose the Bellman EM algorithm (BEM) and the modified Bellman EM algorithm (MBEM) 
by introducing the forward and backward Bellman equations into EM. 
BEM can be more efficient than EM because BEM calculates the forward and backward Bellman equations 
instead of the forward--backward algorithm up to the infinite horizon. 
However, BEM cannot always be more efficient than EM when the size of problems is large 
because BEM calculates an inverse matrix. 
We circumvent this shortcoming in MBEM 
by calculating the forward and backward Bellman equations without the inverse matrix. 
Our numerical experiments demonstrate that the convergence of MBEM is faster than that of EM. }


\usepackage{algorithm,algorithmic}
\usepackage{amsfonts}
\allowdisplaybreaks[4]
\newcommand{\bs}[1]{{\mbox{\boldmath $#1$}}}
\newcommand{\mcal}[1]{\mathcal{#1}}
\newcommand{\mb}[1]{\mathbb{#1}}
\newcommand{\ve}{\varepsilon}

\begin{document}
\section{Introduction}
Markov decision process (MDP) models sequential decision making problems 
and has been used for planning and reinforcement learning \cite{bertsekas2000dynamic,puterman2014markov,sutton1998,sutton2018reinforcement}. 
MDP consists of an environment and an agent. 
The agent observes the state of the environment and controls it by taking actions. 
The planning of MDP is to find the optimal control policy maximizing the objective function, 
which is typically solved by the Bellman equation-based algorithms such as value iteration and policy iteration \cite{bertsekas2000dynamic,puterman2014markov,sutton1998,sutton2018reinforcement}. 

Decentralized partially observable MDP (DEC-POMDP) is an extension of MDP to a multiagent and partially observable setting, 
which models sequential decision making problems by a team of agents \cite{kochenderfer2015decision,oliehoek2010value,Oliehoek2016}. 
DEC-POMDP consists of an environment and multiple agents, and the agents cannot observe the state of the environment and the actions of the other agents completely.  
The agents infer the environmental state and the other agents' actions from their observation histories and control them by taking actions. 
The planning of DEC-POMDP is to find not only the optimal control policy but also the optimal inference policy for each agent, which maximize the objective function \cite{kochenderfer2015decision,oliehoek2010value,Oliehoek2016}. 
Applications of DEC-POMDP include planetary exploration by a team of rovers \cite{Becker2004solving}, target tracking by a team of sensors \cite{Nair2005networked}, and information transmission by a team of devices \cite{Bernstein2002complexity}. 
Since the agents cannot observe the environmental state and the other agents' actions completely, 
it is difficult to extend the Bellman equation-based algorithms for MDP to DEC-POMDP straightforwardly~\cite{bernstein2005,bernstein2009,amato2010a,amato2010b,amato2012}. 

DEC-POMDP can be solved using control as inference \cite{kumar2010,kumar2015a}. 
Control as inference is a framework to interpret a control problem as an inference problem by introducing auxiliary variables \cite{toussaint2006a,todorov2008general,kappen2012optimal,levine2018reinforcement,sun2019tutorial}. 
Although control as inference has several {variants}, 
Toussaint and Storkey showed that the planning of MDP can be interpreted as the maximum likelihood estimation for a latent variable model \cite{toussaint2006a}. 
Thus, the planning of MDP can be solved by EM algorithm, 
which is the typical algorithm for the maximum likelihood estimation of latent variable models \cite{bishop2006pattern}. 
Since the EM algorithm is more general than the Bellman equation-based algorithms, 
it can be straightforwardly extended to POMDP  \cite{toussaint2006b,toussaint2008hierarchical} and DEC-POMDP~\cite{kumar2010,kumar2015a}. 
The computational efficiency of the EM algorithm for DEC-POMDP is comparable to that of other algorithms for DEC-POMDP \cite{kumar2010,kumar2015a,kumar2011,pajarinen2011a,pajarinen2011b}, 
and the extensions to the average reward setting and to the reinforcement learning setting have been studied \cite{pajarinen2013,wu2013monte,liu2016learning}. 

However, the EM algorithm for DEC-POMDP is not efficient enough to be applied to real-world problems, which often have a large number of agents or a large size of an environment. 
Therefore, there are several studies in which improvement of the computational efficiency of the EM algorithm for DEC-POMDP was attempted \cite{kumar2011,pajarinen2011a}. 
Because these studies achieve improvements by restricting possible interactions between agents, their applicability is limited. Therefore, it is desirable to have improvement in the efficiency for more general DEC-POMDP problems.

In order to improve the computational efficiency of EM algorithm for general DEC-POMDP problems, 
there are two problems that need to be resolved. 
The first problem is the forward--backward algorithm up to the infinite horizon.
The EM algorithm for DEC-POMDP uses the forward--backward algorithm, 
which has also been used in EM algorithm for hidden Markov models \cite{bishop2006pattern}. 
However, in the EM algorithm for DEC-POMDP, the forward--backward algorithm needs to be calculated up to the infinite horizon, 
which impairs the computational efficiency \cite{Song2016,Kumar2016}. 
The second problem is the Bellman equation. 
The EM algorithm for DEC-POMDP does not use the Bellman equation, 
which plays a central role in the the planning and in the reinforcement learning for MDP \cite{bertsekas2000dynamic,puterman2014markov,sutton1998,sutton2018reinforcement}. 
Therefore, the EM algorithm for DEC-POMDP cannot use the advanced techniques based on the Bellman equation, 
which makes it possible to solve large-size problems \cite{bertsekas2011approximate,liu2015feature,mnih2015human}. 

\textls[-10]{In some previous studies, resolution of these problems was attempted by replacing the forward--backward algorithm up to the infinite horizon with the Bellman equation \cite{Song2016,Kumar2016}.} 
However, in these studies, the computational efficiency could not be improved completely. 
For example, Song {et al.} replaced the forward--backward algorithm with the Bellman equation and showed that their algorithm is more efficient than EM and other DEC-POMDP algorithms by the numerical experiments \cite{Song2016}. 
However, since a parameter dependency is overlooked in \cite{Song2016}, their algorithm may not find the optimal policy under a general situation (see Appendix \ref{adx: mistake in Song2016} for more details).
Moreover, Kumar {et al.} showed that the forward--backward algorithm can be replaced by linear programming with the Bellman equation as a constraint \cite{Kumar2016}. 
However, their algorithm may be less efficient than the EM algorithm when the size of problems is large. 
Therefore, previous studies have not yet completely improved the computational efficiency of EM algorithm for DEC-POMDP.

In this paper, 
we propose more efficient algorithms for DEC-POMDP than EM algorithm by introducing the forward and backward Bellman equations into it. 
The backward Bellman equation corresponds to the traditional Bellman equation, which has been used in previous studies \cite{Song2016,Kumar2016}. 
In contrast, the forward Bellman equation has not yet been used for the planning of DEC-POMDP explicitly. 
This equation is similar to that recently proposed in the offline reinforcement learning of MDP \cite{hallak2017consistent,gelada2019off,levine2020offline}. 
In the offline reinforcement learning of MDP, the forward Bellman equation is used to correct the difference between the data sampling policy and the policy to be evaluated. 
In the planning of DEC-POMDP, the forward Bellman equation plays the important role in inferring the environmental state. 

We propose the Bellman EM algorithm (BEM) and the modified Bellman EM algorithm (MBEM) by replacing the forward--backward algorithm with the forward and backward Bellman equations.
They are different in terms of how to solve the forward and backward Bellman equations. 
BEM solves the forward and backward Bellman equations by calculating an inverse matrix. 
BEM can be more efficient than EM because BEM does not calculate the forward--backward algorithm up to the infinite horizon. 
However, since BEM calculates the inverse matrix, it cannot always be more efficient than EM when the size of problems is large, 
which is the same problem as \cite{Kumar2016}. 
Actually, BEM is essentially the same as \cite{Kumar2016}. 
In the linear programming problem of \cite{Kumar2016}, the number of variables is equal to that of constraints, which enables us to solve it only from the constraints without the optimization. 
Therefore, the algorithm in {\cite{Kumar2016}} 
 becomes equivalent to BEM, and they suffers from the same problem as~BEM. 

This problem is addressed by MBEM. 
MBEM solves the forward and backward Bellman equations by applying the forward and backward Bellman operators to the arbitrary initial functions infinite times. 
Although MBEM needs to calculate the forward and backward Bellman operators infinite times, which is the same problem with EM, 
MBEM can evaluate approximation errors more tightly owing to the contractibility of these operators. 
It can also utilize the information of the previous iteration owing to the arbitrariness of the initial functions. 
These properties enable MBEM to be more efficient than EM. 
Moreover, MBEM resolves the drawback of BEM because MBEM does not calculate the inverse matrix. 
Therefore, MBEM can be more efficient than EM even when the size of problems is large. 
Our numerical experiments demonstrate that the convergence of MBEM is faster than that of EM regardless of the size of problems. 

The paper is organized as follows: 
In Section~\ref{section: DEC-POMDP}, DEC-POMDP is formulated. 
In Section~\ref{section: EM algorithm for DEC-POMDP}, the EM algorithm for DEC-POMDP, which was proposed in \cite{kumar2010}, is briefly reviewed. 
In Section~\ref{section: Bellman EM algorithm}, the forward and backward Bellman equations are derived, and the Bellman EM algorithm (BEM) is proposed. 
In Section~\ref{section: Modified Bellman EM algorithm}, the forward and backward Bellman operators are defined, and the modified Bellman EM algorithm (MBEM) is proposed. 
In Section~\ref{section: Summary of EM, BEM, and MBEM}, EM, BEM, and MBEM are summarized and compared. 
In Section~\ref{section: Numerical experiment}, the performances of EM, BEM, and MBEM are compared through the numerical experiment. 
In Section~\ref{section: Conclusion}, this paper is concluded, and future works are discussed.

\section{DEC-POMDP}\label{section: DEC-POMDP}

\begin{figure}[H]
	\includegraphics[width=140mm]{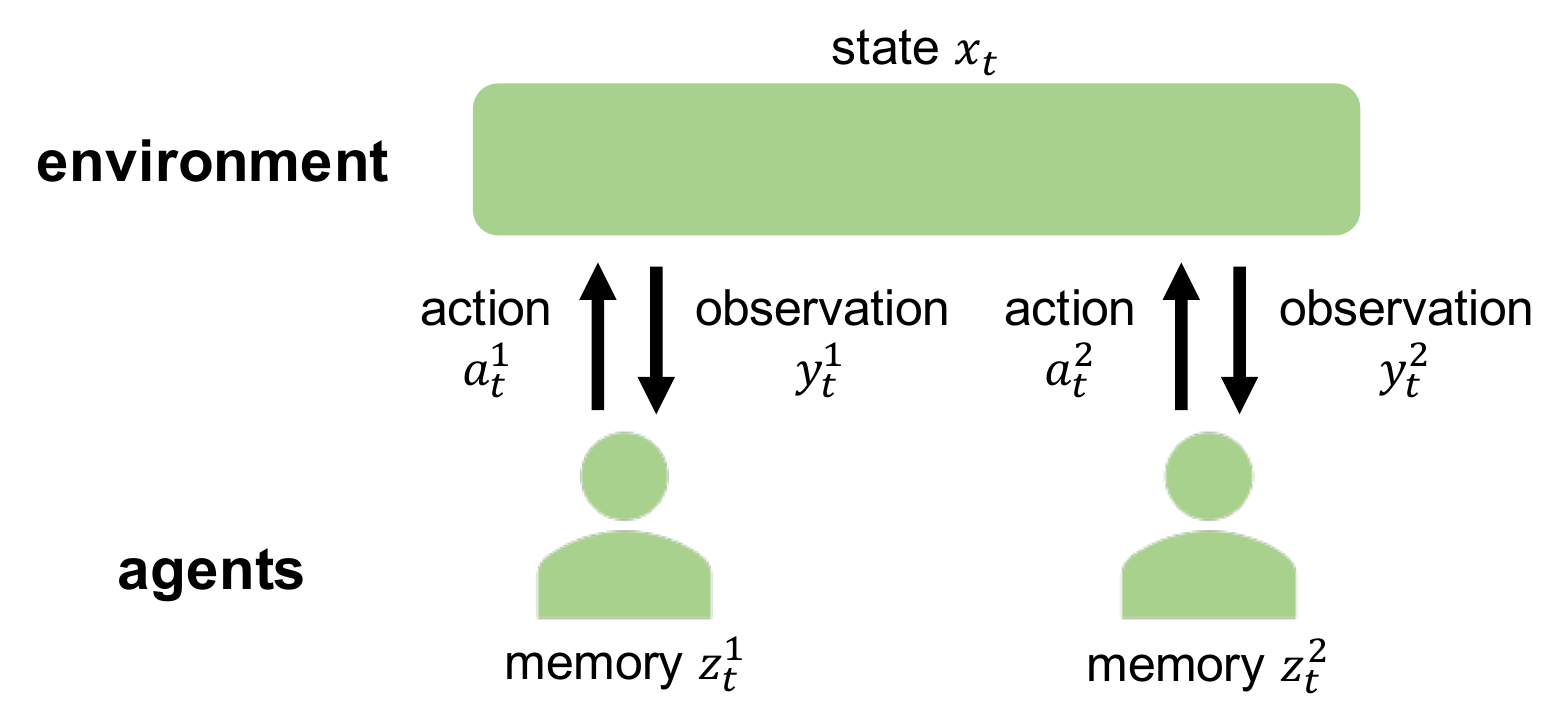}
	\caption{
	Schematic diagram of DEC-POMDP. 
	DEC-POMDP consists of an environment and $N$ agents ($N=2$ in this figure). 
	$x_{t}$ is the state of the environment at time $t$. 
	$y_{t}^{i}$, $z_{t}^{i}$, and $a_{t}^{i}$ are the observation, the memory, and the action available to the agent $i\in\{1,...,N\}$, respectively. 
	The agents update their memories based on their observations, and take their actions based on their memories to control the environmental state. 
	The planning of DEC-POMDP is to find their optimal memory updates and action selections that maximize the objective function.
	}
	\label{fig: model-add}
\end{figure}

\begin{figure}[H]
\widefigure
    	(\textbf{a})
	\begin{minipage}[t][][b]{80mm}
		\includegraphics[width=70mm]{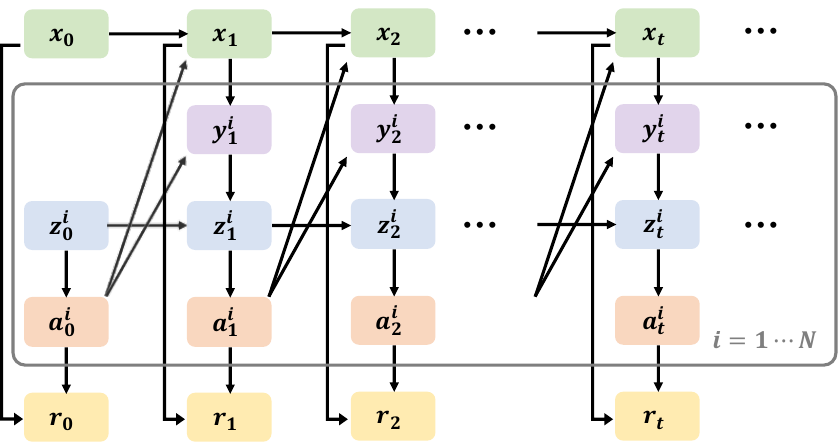}
	\end{minipage}
	(\textbf{b})
	\begin{minipage}[t][][b]{80mm}
		\includegraphics[width=70mm]{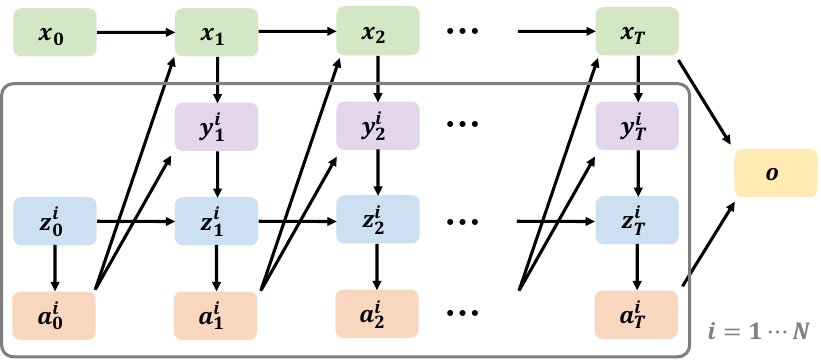}
	\end{minipage}
	\caption{
	Dynamic Bayesian networks of DEC-POMDP (\textbf{a}) and the latent variable model for the time horizon $T\in\{0,1,2,...\}$ (\textbf{b}). 
	$x_{t}$ is the state of the environment at time $t$. 
	$y_{t}^{i}$, $z_{t}^{i}$, and $a_{t}^{i}$ are the observation, the memory, and the action available to the agent $i\in\{1,...,N\}$, respectively. 
	(\textbf{a}) $r_{t}\in\mb{R}$ is the reward, which is generated at each time. 
	(\textbf{b}) $o\in\{0,1\}$ is the optimal variable, which is generated only at the time horizon $T$. 
	}
	\label{fig: model}
\end{figure}

DEC-POMDP consists of an environment and $N$ agents (Figures \ref{fig: model-add} and \ref{fig: model}a) \cite{kumar2010,Oliehoek2016}. 
$x_{t}\in\mcal{X}$ is the state of the environment at time $t$. 
$y_{t}^{i}\in\mcal{Y}^{i}$, $z_{t}^{i}\in\mcal{Z}^{i}$, and $a_{t}^{i}\in\mcal{A}^{i}$ are the observation, the memory, and the action available to the agent $i\in\{1,...,N\}$, respectively. 
$\mcal{X}$, $\mcal{Y}^{i}$, $\mcal{Z}^{i}$, and $\mcal{A}^{i}$ are finite sets. 
$\bs{y}_{t}:=(y_{t}^{1},..,y_{t}^{N})$, $\bs{z}_{t}:=(z_{t}^{1},..,z_{t}^{N})$, and $\bs{a}_{t}:=(a_{t}^{1},..,a_{t}^{N})$ are the joint observation, the joint memory, and the joint action of the $N$ agents, respectively. 

The time evolution of the environmental state $x_{t}$ is given by the initial state probability $p(x_{0})$ and the state transition probability $p(x_{t+1}|x_{t},\bs{a}_{t})$. 
Thus, agents can control the environmental state $x_{t+1}$ by taking appropriate actions $\bs{a}_{t}$. 
The agent $i$ cannot observe the environmental state $x_{t}$ and the joint action $\bs{a}_{t-1}$ completely, and obtains the observation $y_{t}^{i}$ instead of them. 
Thus, the observation $\bs{y}_{t}$ obeys the observation probability $p(\bs{y}_{t}|x_{t},\bs{a}_{t-1})$. 
The agent $i$ updates its memory from $z_{t-1}^{i}$ to $z_{t}^{i}$ based on the observation $y_{t}^{i}$. 
Thus, the memory $z_{t}^{i}$ obeys the initial memory probability $\nu^{i}(z_{0}^{i})$ and the memory transition probability $\lambda^{i}(z_{t}^{i}|z_{t-1}^{i},y_{t}^{i})$. 
The agent $i$ takes the action $a_{t}^{i}$ based on the memory $z_{t}^{i}$ by following the action probability $\pi^{i}(a_{t}^{i}|z_{t}^{i})$. 
The reward function $r(x_{t},\bs{a}_{t})$ defines the amount of reward that is obtained at each step depending on the state of the environment $x_{t}$ and the joint action $\bs{a}_{t}$ taken by the agents.

The objective function in the planning of DEC-POMDP is given by the expected return, which is the expected discounted cumulative reward:
\begin{align}
    J(\bs{\theta})=\mb{E}_{\bs{\theta}}\left[\sum_{t=0}^{\infty}\gamma^{t}r(x_{t},\bs{a}_{t})\right]. 
    \label{eq: expected return}
\end{align}
$\bs{\theta}:=(\bs{\pi},\bs{\lambda},\bs{\nu})$ is the policy, 
where $\bs{\pi}:=(\pi^{1},...,\pi^{N})$, $\bs{\lambda}:=(\lambda^{1},...,\lambda^{N})$, and $\bs{\nu}:=(\nu^{1},...,\nu^{N})$.
$\gamma\in(0,1)$ is the discount factor, which decreases the weight of the future reward. 
The closer $\gamma$ is to 1, the closer the weight of the future reward is to that of the current reward. 

The planning of DEC-POMDP is to find the policy $\bs{\theta}$ that maximizes the expected return $J(\bs{\theta})$ as follows: 
\begin{align}
	\bs{\theta}^{*}:=\arg\max_{\bs{\theta}}J(\bs{\theta}).
	\label{eq: planning in DEC-POMDP}
\end{align}
In other words, the planning of DEC-POMDP is to find how to take the action and how to update the memory for each agent to maximize the expected return. 
\vspace{-6pt}
\end{paracol}
\begin{paracol}{2}
\switchcolumn

\vspace{-6pt}
\section{EM Algorithm for DEC-POMDP}\label{section: EM algorithm for DEC-POMDP}
In this section, we explain the EM algorithm for DEC-POMDP, which was proposed in~\cite{kumar2010}. 

\subsection{Control as Inference}
In this subsection, we show that the planning of DEC-POMDP can be interpreted as the maximum likelihood estimation for a latent variable model (Figure \ref{fig: model}b). 

We introduce two auxiliary random variables: the time horizon $T\in\{0,1,2,...\}$ and the optimal variable $o\in\{0,1\}$. These variables obey the following probabilities: 
\begin{align}
	p(T)&=(1-\gamma)\gamma^{T}\label{eq: time horizon},\\
	p(o=1|x_{T},\bs{a}_{T})&=\bar{r}(x_{T},\bs{a}_{T}):=\frac{r(x_{T},\bs{a}_{T})-r_{\min}}{r_{\max}-r_{\min}}\label{eq: optimal variable}
\end{align}
where $r_{\max}$ and $r_{\min}$ are the maximum and the minimum value of the reward function $r(x,\bs{a})$, respectively. 
Thus, $\bar{r}(x,\bs{a})\in[0,1]$ is satisfied. 

By introducing these variables, DEC-POMDP changes from Figure \ref{fig: model}a to Figure \ref{fig: model}b. 
While Figure \ref{fig: model}a considers the infinite time horizon, Figure \ref{fig: model}b considers the finite time horizon $T$, which obeys Equation (\ref{eq: time horizon}). 
Moreover, while the reward $r_{t}:=r(x_{t},\bs{a}_{t})$ is generated at each time in Figure \ref{fig: model}a, the optimal variable $o$ is generated only at the time horizon $T$ in Figure \ref{fig: model}b.

\begin{Theorem}[\cite{kumar2010}]
\label{th: control as inference}
\rm
The expected return $J(\bs{\theta})$ in DEC-POMDP (Figure \ref{fig: model}a) is linearly related to the likelihood $p(o=1;\bs{\theta})$ in the latent variable model (Figure \ref{fig: model}b) as follows: 
\begin{align}
	J(\bs{\theta})=(1-\gamma)^{-1}\left[(r_{\max}-r_{\min})p(o=1;\bs{\theta})+r_{\min}\right].
	\label{relation between likelihood and expected return}
\end{align}
Note that $o$ is the observable variable, and  $x_{0:T}$, $\bs{y}_{1:T}$, $\bs{z}_{0:T}$, $\bs{a}_{0:T}$, $T$ are the latent variables. 
\end{Theorem}

\begin{proof}
See Appendix \ref{adx: control as inference}. 
\end{proof}

Therefore, the planning of DEC-POMDP is equivalent to the maximum likelihood estimation for the latent variable model as follows: 
\begin{align}
	\bs{\theta}^{*}=\arg\max_{\bs{\theta}}p(o=1;\bs{\theta}).
\end{align}
Intuitively, while the planning of DEC-POMDP is to find the policy which maximizes the reward, 
the maximum likelihood estimation for the latent variable model is to find the policy which maximizes the probability of the optimal variable. 
Since the probability of the optimal variable is proportional to the reward, 
the planning of DEC-POMDP is equivalent to the maximum likelihood estimation for the latent variable model. 

\subsection{EM Algorithm}
Since the planning of DEC-POMDP can be interpreted as the maximum likelihood estimation for the latent variable model, it can be solved by the EM algorithm \cite{kumar2010}. 
EM algorithm is the typical algorithm for the maximum likelihood estimation of latent variable models, which iterates two steps, E step and M step \cite{bishop2006pattern}. 

In the E step, we calculate the Q function, which is defined as follows: 
\begin{align}
	Q(\bs{\theta};\bs{\theta}_{k}):=\mb{E}_{\bs{\theta}_{k}}\left[\left.\log p(o=1,x_{0:T},\bs{y}_{0:T},\bs{z}_{0:T},\bs{a}_{0:T},T;\bs{\theta})\right|o=1\right]
	\label{Eq: Q-function}
\end{align}
where $\bs{\theta}_{k}$ is the current estimator of the optimal policy. 

In the M step, we update $\bs{\theta}_{k}$ to $\bs{\theta}_{k+1}$ by maximizing the Q function as follows: 
\begin{align}
	\bs{\theta}_{k+1}:=\arg\max_{\bs{\theta}}Q(\bs{\theta};\bs{\theta}_{k}).
	\label{Eq: M step}
\end{align}

Since each iteration between the E step and the M step monotonically increases the likelihood $p(o=1;\bs{\theta}_{k})$, 
we can find $\bs{\theta}^{*}$ that locally maximizes the likelihood $p(o=1;\bs{\theta})$. 

\subsection{M Step}

\begin{Proposition}[\cite{kumar2010}]
\label{th: M step}
\rm
In the EM algorithm for DEC-POMDP, Equation (\ref{Eq: M step}) can be calculated as~follows:
{\small
\begin{align}
	&\pi_{k+1}^{i}(a|z)
	=\frac{\sum_{\bs{a}^{-i},\bs{z}^{-i}}\bs{\pi}_{k}(\bs{a}|\bs{z})\sum_{x,x',\bs{z}'}p(x',\bs{z}'|x,\bs{z},\bs{a};\bs{\lambda}_{k})F(x,\bs{z};\bs{\theta}_{k})\left(\bar{r}(x,\bs{a})+\gamma V(x,\bs{z};\bs{\theta}_{k})\right)}{\sum_{\bs{a},\bs{z}}\bs{\pi}_{k}(\bs{a}|\bs{z})\sum_{x,x',\bs{z}'}p(x',\bs{z}'|x,\bs{z},\bs{a};\bs{\lambda}_{k})F(x,\bs{z};\bs{\theta}_{k})\left(\bar{r}(x,\bs{a})+\gamma V(x,\bs{z};\bs{\theta}_{k})\right)}
	\label{Eq: M step-pi},\\
	&\lambda_{k+1}^{i}(z'|z,y')
	=\frac{\sum_{\bs{z}^{-i'},\bs{z}^{-i},\bs{y}^{-i'}}\bs{\lambda}_{k}(\bs{z}^{i'}|\bs{z}^{i},\bs{y}^{i'})\sum_{x',x}p(x',\bs{y}'|x,\bs{z};\bs{\pi}_{k})F(x,\bs{z};\bs{\theta}_{k}) V(x',\bs{z}';\bs{\theta}_{k})}{\sum_{\bs{z}^{'},\bs{z},\bs{y}^{'}}\bs{\lambda}_{k}(\bs{z}^{i'}|\bs{z}^{i},\bs{y}^{i'})\sum_{x',x}p(x',\bs{y}'|x,\bs{z};\bs{\pi}_{k})F(x,\bs{z};\bs{\theta}_{k}) V(x',\bs{z}';\bs{\theta}_{k})},
	\label{Eq: M step-lambda}\\
	&\nu_{k+1}^{i}(z)=\frac{\sum_{\bs{z}^{-i}}\bs{\nu}_{k}(\bs{z})\sum_{x}p_{0}(x)V(x,\bs{z};\bs{\theta}_{k})}{\sum_{\bs{z}}\bs{\nu}_{k}(\bs{z})\sum_{x}p_{0}(x)V(x,\bs{z};\bs{\theta}_{k})}.
	\label{Eq: M step-nu}
\end{align}
}$\bs{a}^{-i}:=(a^{1},...,a^{i-1},a^{i+1},...,a^{N})$. 
$\bs{y}^{-i}$ and $\bs{z}^{-i}$ are defined in the same way.  
$F(x,\bs{z};\bs{\theta})$ and $V(x,\bs{z};\bs{\theta})$ are defined as follows:
\begin{align}
	F(x,\bs{z};\bs{\theta})&:=\sum_{t=0}^{\infty}\gamma^{t}p_{t}(x,\bs{z};\bs{\theta})\label{eq: forward var},\\
	V(x,\bs{z};\bs{\theta})&:=\sum_{t=0}^{\infty}\gamma^{t}p_{0}(o=1|x,\bs{z},T=t;\bs{\theta})\label{eq: backward var}
\end{align}
where $p_{t}(x,\bs{z};\bs{\theta}):=p(x_{t}=x,\bs{z}_{t}=\bs{z};\bs{\theta})$, 
and $p_{0}(o=1|x,\bs{z},T;\bs{\theta}):=p(o=1|x_{0}=x,\bs{z}_{0}=\bs{z},T;\bs{\theta})$. 
\end{Proposition}

\begin{proof}
See Appendix \ref{adx: M step}. 
\end{proof}

$F(x,\bs{z};\bs{\theta})$ quantifies the frequency of the state $x$ and the memory $\bs{z}$, 
which is called the frequency function in this paper.
$V(x,\bs{z};\bs{\theta})$ quantifies the probability of $o=1$ 
when the initial state and memory are $x$ and $\bs{z}$, respectively. 
Since the probability of $o=1$ is proportional to the reward, 
$V(x,\bs{z};\bs{\theta})$ is called the value function in this paper. 
Actually, $V(x,\bs{z};\bs{\theta})$ corresponds to the value function \cite{Kumar2016}.

\subsection{E Step}

$F(x,\bs{z};\bs{\theta}_{k})$ and $V(x,\bs{z};\bs{\theta}_{k})$ need to be obtained to calculate Equations (\ref{Eq: M step-pi})--(\ref{Eq: M step-nu}). 
In \cite{kumar2010}, $F(x,\bs{z};\bs{\theta}_{k})$ and $V(x,\bs{z};\bs{\theta}_{k})$ are calculated by the forward--backward algorithm, 
which has been used in EM algorithm for the hidden Markov model \cite{bishop2006pattern}. 

In \cite{kumar2010}, the forward probability $\alpha_{t}(x,\bs{z})$ and the backward probability $\beta_{t}(x,\bs{z})$ are defined as follows: 
\begin{align}
	\alpha_{t}(x,\bs{z};\bs{\theta}_{k})&:=p_{t}(x,\bs{z};\bs{\theta}_{k}),\\
	\beta_{t}(x,\bs{z};\bs{\theta}_{k})&:=p_{0}(o=1|x,\bs{z},T=t;\bs{\theta}_{k}).
\end{align}
It is easy to calculate $\alpha_{0}(x,\bs{z};\bs{\theta}_{k})$ and $\beta_{0}(x,\bs{z};\bs{\theta}_{k})$ as follows: 
\begin{align}
	\alpha_{0}(x,\bs{z};\bs{\theta}_{k})&=p_{0}(x,\bs{z};\bs{\nu}_{k}):=p_{0}(x)\bs{\nu}_{k}(\bs{z}),\label{eq: alpha0}\\
	\beta_{0}(x,\bs{z};\bs{\theta}_{k})&=\bar{r}(x,\bs{z};\bs{\pi}_{k}):=\sum_{\bs{a}}\bs{\pi}_{k}(\bs{a}|\bs{z})\bar{r}(x,\bs{a}).\label{eq: beta0}
\end{align}
Moreover, $\alpha_{t+1}(x,\bs{z};\bs{\theta}_{k})$ and $\beta_{t+1}(x,\bs{z};\bs{\theta}_{k})$ are easily calculated from $\alpha_{t}(x,\bs{z};\bs{\theta}_{k})$ and $\beta_{t}(x,\bs{z};\bs{\theta}_{k})$: 
\begin{align}
	\alpha_{t+1}(x,\bs{z};\bs{\theta}_{k})&=\sum_{x',\bs{z}'}p(x,\bs{z}|x',\bs{z}';\bs{\theta}_{k})\alpha_{t}(x',\bs{z}';\bs{\theta}_{k})\label{eq: forward eq},\\
	\beta_{t+1}(x,\bs{z};\bs{\theta}_{k})&=\sum_{x',\bs{z}'}\beta_{t}(x',\bs{z}';\bs{\theta}_{k})p(x',\bs{z}'|x,\bs{z};\bs{\theta}_{k})\label{eq: backward eq}
\end{align}
where 
\begin{align}
    &p(x',\bs{z}'|x,\bs{z};\bs{\theta}_{k})
    =\sum_{\bs{y}',\bs{a}}\bs{\lambda}_{k}(\bs{z}'|\bs{z},\bs{y}')p(\bs{y}'|x',\bs{a})p(x'|x,\bs{a})\bs{\pi}_{k}(\bs{a}|\bs{z}).
    \label{eq: p(x,z|x,z)}
\end{align}
Equations (\ref{eq: forward eq}) and (\ref{eq: backward eq}) are called the forward and backward equations, respectively. 
Using Equations (\ref{eq: alpha0})--(\ref{eq: backward eq}), 
$\alpha_{t}(x,\bs{z};\bs{\theta}_{k})$ and $\beta_{t}(x,\bs{z};\bs{\theta}_{k})$ can be efficiently calculated from $t=0$ to $t=\infty$, 
which is called the forward--backward algorithm \cite{bishop2006pattern}. 

By calculating the forward--backward algorithm from $t=0$ to $t=\infty$, 
$F(x,\bs{z};\bs{\theta}_{k})$ and $V(x,\bs{z};\bs{\theta}_{k})$ can be obtained as follows \cite{kumar2010}: 
\begin{align}
	F(x,\bs{z};\bs{\theta}_{k})&=\sum_{t=0}^{\infty}\gamma^{t}\alpha_{t}(x,\bs{z};\bs{\theta}_{k}),\\
	V(x,\bs{z};\bs{\theta}_{k})&=\sum_{t=0}^{\infty}\gamma^{t}\beta_{t}(x,\bs{z};\bs{\theta}_{k}).
\end{align}
However, $F(x,\bs{z};\bs{\theta}_{k})$ and $V(x,\bs{z};\bs{\theta}_{k})$ cannot be calculated exactly by this approach 
because it is practically impossible to calculate the forward--backward algorithm until $t=\infty$. 
Therefore, the forward--backward algorithm needs to be terminated at $t=T_{\max}$, where $T_{\max}$ is finite. In this case, $F(x,\bs{z};\bs{\theta}_{k})$ and $V(x,\bs{z};\bs{\theta}_{k})$ are approximated as follows: 
\begin{align}
	F(x,\bs{z};\bs{\theta}_{k})&=\sum_{t=0}^{\infty}\gamma^{t}\alpha_{t}(x,\bs{z};\bs{\theta}_{k})\approx\sum_{t=0}^{T_{\max}}\gamma^{t}\alpha_{t}(x,\bs{z};\bs{\theta}_{k}),\label{eq: F of EM}\\
	V(x,\bs{z};\bs{\theta}_{k})&=\sum_{t=0}^{\infty}\gamma^{t}\beta_{t}(x,\bs{z};\bs{\theta}_{k})\approx\sum_{t=0}^{T_{\max}}\gamma^{t}\beta_{t}(x,\bs{z};\bs{\theta}_{k}).\label{eq: V of EM}
\end{align}
$T_{\max}$ needs to be large enough to reduce the approximation errors. 
In the previous study, a heuristic termination condition was proposed as follows
\cite{kumar2010}: 
\begin{align}
	\gamma^{T_{\max}}p(o=1|T_{\max};\bs{\theta}_{k})
	\ll\sum_{T=0}^{T_{\max}-1}\gamma^{T}p(o=1|T;\bs{\theta}_{k}).
	\label{eq: heuristic termination condition}
\end{align}
$p(o=1|T;\bs{\theta}_{k})=\sum_{x,\bs{z}}\alpha_{T'}(x,\bs{z};\bs{\theta}_{k})\beta_{T''}(x,\bs{z};\bs{\theta}_{k})$ where $T=T'+T''$. 
However, the relation between $T_{\max}$ and the approximation errors is unclear in Equation (\ref{eq: heuristic termination condition}). 
We propose a new termination condition to guarantee the approximation errors as follows: 

\begin{Proposition}\label{prop: Tmax}
\rm
We set an acceptable error bound $\varepsilon>0$. 
If
\begin{align}
    T_{\max}>\frac{\log(1-\gamma)\varepsilon}{\log\gamma}-1
    \label{eq: Tmax}
\end{align}
is satisfied, then
\begin{align}
    \left\|F(x,\bs{z};\bs{\theta}_{k})-\sum_{t=0}^{T_{\max}}\gamma^{t}\alpha_{t}(x,\bs{z};\bs{\theta}_{k})\right\|_{\infty}&<\varepsilon\label{eq: error of F in EM},\\
    \left\|V(x,\bs{z};\bs{\theta}_{k})-\sum_{t=0}^{T_{\max}}\gamma^{t}\beta_{t}(x,\bs{z};\bs{\theta}_{k})\right\|_{\infty}&<\varepsilon\label{eq: error of V in EM}
\end{align}
are satisfied. 
\end{Proposition}

\begin{proof}
See Appendix \ref{adx: Tmax}. 
\end{proof}
\newpage
\subsection{Summary}

In summary, the EM algorithm for DEC-POMDP is given by Algorithm \ref{alg: EM}. In the E step, we calculate $\alpha_{t}(x,\bs{z};\bs{\theta}_{k})$ and $\beta_{t}(x,\bs{z};\bs{\theta}_{k})$ from $t=0$ to $t=T_{\max}$ by the forward--backward algorithm. In M step, we update $\bs{\theta}_{k}$ to $\bs{\theta}_{k+1}$ using Equations (\ref{Eq: M step-pi})--(\ref{Eq: M step-nu}). 
The time complexities of the E step and the M step are $\mcal{O}((|\mcal{X}||\mcal{Z}|)^2T_{\max})$ and $\mcal{O}((|\mcal{X}||\mcal{Z}|)^2|\mcal{Y}||\mcal{A}|)$, respectively. Note that $\mcal{A}:=\otimes_{i=1}^{N}\mcal{A}^{i}$, and $\mcal{Y}$ and $\mcal{Z}$ are defined in the same way. 
The EM algorithm for DEC-POMDP is less efficient when the discount factor $\gamma$ is closer to 1 or the acceptable error bound $\ve$ is smaller 
because $T_{\max}$ needs to be larger in these cases. 
\begin{algorithm}[H]
\caption{EM algorithm for DEC-POMDP}
\label{alg: EM}
\begin{algorithmic}
{\small
\STATE $k\leftarrow0$, Initialize $\bs{\theta}_{k}$. 
\STATE $T_{\max}\leftarrow\lceil(\log(1-\gamma)\varepsilon)/\log\gamma-1\rceil$
\WHILE{$\bs{\theta}_{k}$ or $J(\bs{\theta}_{k})$ do not converge}
\STATE Calculate $p(x',\bs{z}'|x,\bs{z};\bs{\theta_{k}})$ by Equation (\ref{eq: p(x,z|x,z)}).
\STATE //---E step---//
\STATE $\alpha_{0}(x,\bs{z};\bs{\theta}_{k})\leftarrow p_{0}(x,\bs{z};\bs{\nu}_{k})$
\STATE $\beta_{0}(x,\bs{z};\bs{\theta}_{k})\leftarrow\bar{r}(x,\bs{z};\bs{\pi}_{k})$
\FOR{$t=1,2,...,T_{\max}$}
\STATE $\alpha_{t}(x,\bs{z};\bs{\theta}_{k})\leftarrow\sum_{x',\bs{z}'}p(x,\bs{z}|x',\bs{z}';\bs{\theta}_{k})\alpha_{t-1}(x',\bs{z}';\bs{\theta}_{k})$
\STATE $\beta_{t}(x,\bs{z};\bs{\theta}_{k})\leftarrow\sum_{x',\bs{z}'}\beta_{t-1}(x',\bs{z}';\bs{\theta}_{k})p(x',\bs{z}'|x,\bs{z};\bs{\theta}_{k})$
\ENDFOR
\STATE $F(x,\bs{z};\bs{\theta}_{k})\leftarrow\sum_{t=0}^{T_{\max}}\gamma^{t}\alpha_{t}(x,\bs{z};\bs{\theta}_{k})$
\STATE $V(x,\bs{z};\bs{\theta}_{k})\leftarrow\sum_{t=0}^{T_{\max}}\gamma^{t}\beta_{t}(x,\bs{z};\bs{\theta}_{k})$
\STATE //---M step---//
\STATE Update $\bs{\theta}_{k}$ to $\bs{\theta}_{k+1}$ by Equations (\ref{Eq: M step-pi})--(\ref{Eq: M step-nu}).
\STATE $k\leftarrow k+1$
\ENDWHILE
\RETURN $\bs{\theta}_{k}$
}
\end{algorithmic}
\end{algorithm}

\vspace{-6pt}
\section{Bellman EM Algorithm}\label{section: Bellman EM algorithm}
In the EM algorithm for DEC-POMDP, $\alpha_{t}(x,\bs{z};\bs{\theta}_{k})$ and $\beta_{t}(x,\bs{z};\bs{\theta}_{k})$ are calculated from $t=0$ to $t=T_{\max}$ to obtain $F(x,\bs{z};\bs{\theta}_{k})$ and $V(x,\bs{z};\bs{\theta}_{k})$. 
However, $T_{\max}$ needs to be large to reduce the approximation errors of $F(x,\bs{z};\bs{\theta}_{k})$ and $V(x,\bs{z};\bs{\theta}_{k})$, which impairs the computational efficiency of the EM algorithm for DEC-POMDP \cite{Song2016,Kumar2016}. 
In this section, we calculate $F(x,\bs{z};\bs{\theta}_{k})$ and $V(x,\bs{z};\bs{\theta}_{k})$ directly without calculating $\alpha_{t}(x,\bs{z};\bs{\theta}_{k})$ and $\beta_{t}(x,\bs{z};\bs{\theta}_{k})$ to resolve the drawback of EM.

\subsection{Forward and Backward Bellman Equations}
The following equations are useful to obtain $F(x,\bs{z};\bs{\theta}_{k})$ and $V(x,\bs{z};\bs{\theta}_{k})$ directly: 

\begin{Theorem}\label{th: bellman equations}
\rm
$F(x,\bs{z};\bs{\theta})$ and $V(x,\bs{z};\bs{\theta})$ satisfy the following equations:
\begin{align}
	F(x,\bs{z};\bs{\theta})=&p_{0}(x,\bs{z};\bs{\nu})+\gamma\sum_{x',\bs{z}'}p(x,\bs{z}|x',\bs{z}';\bs{\theta})F(x',\bs{z}';\bs{\theta})\label{eq: FBE},\\
	V(x,\bs{z};\bs{\theta})=&\bar{r}(x,\bs{z};\bs{\pi})+\gamma\sum_{x',\bs{z}'}p(x',\bs{z}'|x,\bs{z};\bs{\theta})V(\bs{x}',\bs{z}';\bs{\theta})\label{eq: BBE}.
\end{align}
Equations (\ref{eq: FBE}) and (\ref{eq: BBE}) are called the forward Bellman equation and the backward Bellman equation, respectively. 
\end{Theorem}

\begin{proof}
See Appendix \ref{adx: bellman equations}. 
\end{proof}

Note that the direction of time is different between Equations (\ref{eq: FBE}) and (\ref{eq: BBE}). 
In Equation~(\ref{eq: FBE}), $x'$ and $\bs{z}'$ are earlier state and memory than $x$ and $\bs{z}$, respectively. 
In Equation (\ref{eq: BBE}), $x'$ and $\bs{z}'$ are later state and memory than $x$ and $\bs{z}$, respectively. 

The backward Bellman Equation (\ref{eq: BBE}) corresponds to the traditional Bellman equation, 
which has been used in other algorithms for DEC-POMDP \cite{bernstein2005,bernstein2009,amato2010a,amato2010b}. 
In contrast, the forward Bellman equation, which is introduced in this paper, is similar to that recently proposed in the offline reinforcement learning \cite{hallak2017consistent,gelada2019off,levine2020offline}.

Since the forward and backward Bellman equations are linear equations, they can be solved exactly as follows:
\begin{align}
	\bs{F}(\bs{\theta})&=(\bs{I}-\gamma\bs{P}(\bs{\theta}))^{-1}\bs{p}(\bs{\nu}),\label{eq: solution of FBE}\\
	\bs{V}(\bs{\theta})&=((\bs{I}-\gamma\bs{P}(\bs{\theta}))^{-1})^{T}\bs{r}(\bs{\pi})\label{eq: solution of BBE}
\end{align}
where 
\begin{align}
	&F_{i}(\bs{\theta}):=F((x,\bs{z})=i;\bs{\theta}), V_{i}(\bs{\theta}):=V((x,\bs{z})=i;\bs{\theta})\nonumber\\
	&P_{ij}(\bs{\theta}):=p((x',\bs{z}')=i|(x,\bs{z})=j;\bs{\theta})\nonumber\\
	&p_{i}(\bs{\nu}):=p_{0}((x,\bs{z})=i;\bs{\nu}), r_{i}(\bs{\pi}):=\bar{r}((x,\bs{z})=i;\bs{\pi})\nonumber
\end{align}
Therefore, we can obtain $F(x,\bs{z};\bs{\theta}_{k})$ and $V(x,\bs{z};\bs{\theta}_{k})$ from the forward and backward Bellman~equations.

\subsection{Bellman EM Algorithm (BEM)}
The forward--backward algorithm from $t=0$ to $t=T_{\max}$ in the EM algorithm for DEC-POMDP can be replaced by the forward and backward Bellman equations. 
In this paper, the EM algorithm for DEC-POMDP that uses the forward and backward Bellman equations instead of the forward--backward algorithm from $t=0$ to $t=T_{\max}$ is called the Bellman EM algorithm (BEM). 

\subsection{Comparison of EM and BEM}
BEM is summarized as Algorithm \ref{alg: BEM}. 
The M step in Algorithm \ref{alg: BEM} is almost the same as that in Algorithm \ref{alg: EM}---only the E step is different. 
While the time complexity of the E step in EM is $\mcal{O}((|\mcal{X}||\mcal{Z}|)^2T_{\max})$, that in BEM is $\mcal{O}((|\mcal{X}||\mcal{Z}|)^3)$. 
\begin{algorithm}[H]
\caption{Bellman EM algorithm (BEM)}
\label{alg: BEM}
\begin{algorithmic}
\STATE $k\leftarrow0$, Initialize $\bs{\theta}_{k}$. 
\WHILE{$\bs{\theta}_{k}$ or $J(\bs{\theta}_{k})$ do not converge}
\STATE Calculate $p(x',\bs{z}'|x,\bs{z};\bs{\theta_{k}})$ by Equation (\ref{eq: p(x,z|x,z)}). 
\STATE //---E step---//
\STATE $\bs{F}(\bs{\theta}_{k})\leftarrow(\bs{I}-\gamma\bs{P}(\bs{\theta}_{k}))^{-1}\bs{p}(\bs{\nu}_{k})$
\STATE $\bs{V}(\bs{\theta}_{k})\leftarrow((\bs{I}-\gamma\bs{P}(\bs{\theta}_{k}))^{-1})^{T}\bs{r}(\bs{\pi}_{k})$
\STATE //---M step---//
\STATE Update $\bs{\theta}_{k}$ to $\bs{\theta}_{k+1}$ by Equations (\ref{Eq: M step-pi})--(\ref{Eq: M step-nu}).
\STATE $k\leftarrow k+1$
\ENDWHILE
\RETURN $\bs{\theta}_{k}$
\end{algorithmic}
\end{algorithm}

BEM can calculate $F(x,\bs{z};\bs{\theta}_{k})$ and $V(x,\bs{z};\bs{\theta}_{k})$ exactly. 
Moreover, BEM can be more efficient than EM when the discount factor $\gamma$ is close to 1 or the acceptable error bound $\ve$ is small 
because $T_{\max}$ needs to be large enough in these cases. 
However, when the size of the state space $|\mcal{X}|$ or that of the joint memory space $|\mcal{Z}|$ is large, 
BEM cannot always be more efficient than EM because BEM needs to calculate the inverse matrix $(\bs{I}-\gamma\bs{P}(\bs{\theta}_{k}))^{-1}$. 
To circumvent this shortcoming, we propose a new algorithm, the modified Bellman EM algorithm (MBEM), to obtain $F(x,\bs{z};\bs{\theta}_{k})$ and $V(x,\bs{z};\bs{\theta}_{k})$ without calculating the inverse~matrix.

\section{Modified Bellman EM Algorithm}\label{section: Modified Bellman EM algorithm}
\subsection{Forward and Backward Bellman Operators}
We define the forward and backward Bellman operators as follows: 
\begin{align}
	A_{\bs{\theta}}f(x,\bs{z}):=&p_{0}(x,\bs{z};\bs{\nu})
	+\gamma\sum_{x',\bs{z}'}p(x,\bs{z}|x',\bs{z}';\bs{\theta})f(x',\bs{z}'),\\
	B_{\bs{\theta}}v(x,\bs{z}):=&\bar{r}(x,\bs{z};\bs{\pi})
	+\gamma\sum_{x',\bs{z}'}p(x',\bs{z}'|x,\bs{z};\bs{\theta})v(\bs{x}',\bs{z}')
\end{align}
where $^\forall f, v:\mcal{X}\times\mcal{Z}\to\mb{R}$. 
From the forward and backward Bellman equations, $A_{\bs{\theta}}$ and $B_{\bs{\theta}}$ satisfy the following equations: 
\begin{align}
    F(x,\bs{z};\bs{\theta})=A_{\bs{\theta}}F(x,\bs{z};\bs{\theta}),\label{eq: the fixed point of A}\\
    V(x,\bs{z};\bs{\theta})=B_{\bs{\theta}}V(x,\bs{z};\bs{\theta}).\label{eq: the fixed point of B}
\end{align}
Thus, $F(x,\bs{z};\bs{\theta}_{k})$ and $V(x,\bs{z};\bs{\theta}_{k})$ are the fixed points of $A_{\bs{\theta}}$ and $B_{\bs{\theta}}$, respectively. 
$A_{\bs{\theta}}$ and $B_{\bs{\theta}}$ have the following useful property: 

\begin{Proposition}\label{prop: contractibility of Bellman operators}
\rm
$A_{\bs{\theta}}$ and $B_{\bs{\theta}}$ are contractive operators as follows:
\begin{align}
	&\|A_{\bs{\theta}}f-A_{\bs{\theta}}g\|_{1}\leq\gamma\|f-g\|_{1}\label{eq: contractibility of A},\\
	&\|B_{\bs{\theta}}u-B_{\bs{\theta}}v\|_{\infty}\leq\gamma\|u-v\|_{\infty}\label{eq: contractibility of B}
\end{align}
where $^\forall f, g, u, v:\mcal{X}\times\mcal{Z}\to\mb{R}$. 
\end{Proposition}

\begin{proof}
See Appendix \ref{adx: contractibility of Bellman operators}.
\end{proof}

Note that the norm is different between Equations (\ref{eq: contractibility of A}) and (\ref{eq: contractibility of B}). It is caused by the difference of the time direction between $A_{\bs{\theta}}$ and $B_{\bs{\theta}}$. 
While $x'$ and $\bs{z}'$ are earlier state and memory than $x$ and $\bs{z}$, respectively, in the forward Bellman operator $A_{\bs{\theta}}$, 
$x'$ and $\bs{z}'$ are later state and memory than $x$ and $\bs{z}$, respectively, in the backward Bellman operator $B_{\bs{\theta}}$. 

We obtain $F(x,\bs{z};\bs{\theta}_{k})$ and $V(x,\bs{z};\bs{\theta}_{k})$ using Equations (\ref{eq: the fixed point of A})--(\ref{eq: contractibility of B}), 
 as follows: 

\begin{Proposition}\label{prop: F and V from Bellman operators}
\rm
\begin{align}
	\lim_{L\to\infty}A_{\bs{\theta}}^{L}f(x,\bs{z})&=F(x,\bs{z};\bs{\theta}),
	\label{eq: FV derived from BO}\\
	\lim_{L\to\infty}B_{\bs{\theta}}^{L}v(x,\bs{z})&=V(x,\bs{z};\bs{\theta})
	\label{eq: BV derived from BO}
\end{align}
where $^\forall f, v:\mcal{X}\times\mcal{Z}\to\mb{R}$.
\end{Proposition}

\begin{proof}
See Appendix \ref{adx: F and V from Bellman operators}.
\end{proof}

Therefore, it is shown that $F(x,\bs{z};\bs{\theta}_{k})$ and $V(x,\bs{z};\bs{\theta}_{k})$ can be calculated by applying the forward and backward Bellman operators, $A_{\bs{\theta}}$ and $B_{\bs{\theta}}$, 
to arbitrary initial functions, $f(x,\bs{z})$ and $v(x,\bs{z})$, infinite times. 

\subsection{Modified Bellman EM Algorithm (MBEM)}
The calculation of the forward and backward Bellman equations in BEM can be replaced by that of the forward and backward Bellman operators. 
In this paper, BEM that uses the forward and backward Bellman operators instead of the forward and backward Bellman equations is called modified Bellman EM algorithm (MBEM). 

\subsection{Comparison of EM, BEM, and MBEM}
Since MBEM does not need the inverse matrix, 
MBEM can be more efficient than BEM when the size of the state space $|\mcal{X}|$ and that of the joint memory space $|\mcal{Z}|$ are large. 
Thus, MBEM resolves the drawback of BEM. 

On the other hand, MBEM has the same problem as EM. 
MBEM calculates $A_{\bs{\theta}_{k}}$ and $B_{\bs{\theta}_{k}}$ infinite times to obtain $F(x,\bs{z};\bs{\theta}_{k})$ and $V(x,\bs{z};\bs{\theta}_{k})$. 
However, since it is practically impossible to calculate $A_{\bs{\theta}_{k}}$ and $B_{\bs{\theta}_{k}}$ infinite times, 
the calculation of $A_{\bs{\theta}_{k}}$ and $B_{\bs{\theta}_{k}}$ needs to be terminated after $L_{\max}$ times, where $L_{\max}$ is finite. 
In this case, $F(x,\bs{z};\bs{\theta}_{k})$ and $V(x,\bs{z};\bs{\theta}_{k})$ are approximated as follows: 
\begin{align}
	F(x,\bs{z};\bs{\theta}_{k})=&A_{\bs{\theta}_{k}}^{\infty}f(x,\bs{z})\approx A_{\bs{\theta}_{k}}^{L_{\max}}f(x,\bs{z}),
	\label{eq: F of MBEM}\\
	V(x,\bs{z};\bs{\theta}_{k})=&B_{\bs{\theta}_{k}}^{\infty}v(x,\bs{z})\approx B_{\bs{\theta}_{k}}^{L_{\max}}v(x,\bs{z}).
	\label{eq: V of MBEM}
\end{align}
$L_{\max}$ needs to be large enough to reduce the approximation errors of $F(x,\bs{z};\bs{\theta}_{k})$ and $V(x,\bs{z};\bs{\theta}_{k})$, 
which impairs the computational efficiency of MBEM. 
Thus, MBEM can potentially suffer from the same problem as EM. 
However, we can theoretically show that MBEM is more efficient than EM by comparing $T_{\max}$ and $L_{\max}$ 
under the condition that the approximation errors of $F(x,\bs{z};\bs{\theta}_{k})$ and $V(x,\bs{z};\bs{\theta}_{k})$ are smaller than the acceptable error bound $\ve$.

When $f(x,\bs{z})=p_{0}(x,\bs{z};\bs{\nu}_{k})$ and $v(x,\bs{z})=\bar{r}(x,\bs{z};\bs{\pi}_{k})$, 
Equations (\ref{eq: F of MBEM}) and (\ref{eq: V of MBEM}) can be calculated as follows: 
\begin{align}
	A_{\bs{\theta}_{k}}^{L_{\max}}f(x,\bs{z})&=\sum_{t=0}^{L_{\max}}\gamma^{t}p_{t}(x,\bs{z};\bs{\theta}),\\
	B_{\bs{\theta}_{k}}^{L_{\max}}v(x,\bs{z})&=\sum_{t=0}^{L_{\max}}\gamma^{t}p_{0}(o=1|x,\bs{z},T=t;\bs{\theta})
\end{align}
\textls[-15]{which are the same with Equations (\ref{eq: F of EM}) and (\ref{eq: V of EM}), respectively. 
Thus, in this case, $L_{\max}=T_{\max}$,} and  the computational efficiency of MBEM is the same as that of EM. 
However, MBEM has two useful properties that EM does not have, 
and therefore, MBEM can be more efficient than EM. 
In the following, we explain these properties in more detail. 

The first property of MBEM is the contractibility of the forward and backward Bellman operators, $A_{\bs{\theta}_{k}}$ and $B_{\bs{\theta}_{k}}$. 
From the contractibility of the Bellman operators, $L_{\max}$ is determined adaptively as follows: 
\begin{Proposition}\label{prop: the termination condition by the contractibility of the Bellman operators}
\rm
We set an acceptable error bound $\varepsilon>0$. 
If
\begin{align}
    &\left\|A_{\bs{\theta}_{k}}^{L}f-A_{\bs{\theta}_{k}}^{L-1}f\right\|_{1}<\frac{1-\gamma}{\gamma}\varepsilon,\\
    &\left\|B_{\bs{\theta}_{k}}^{L}v-B_{\bs{\theta}_{k}}^{L-1}v\right\|_{\infty}<\frac{1-\gamma}{\gamma}\varepsilon
\end{align}
are satisfied, then
\begin{align}
    \left\|F(x,\bs{z};\bs{\theta}_{k})-A_{\bs{\theta}_{k}}^{L}f(x,\bs{z};\bs{\theta}_{k})\right\|_{\infty}&<\varepsilon\label{eq: approx error of MBEM-F},\\
    \left\|V(x,\bs{z};\bs{\theta}_{k})-B_{\bs{\theta}_{k}}^{L}v(x,\bs{z};\bs{\theta}_{k})\right\|_{\infty}&<\varepsilon\label{eq: approx error of MBEM-B}
\end{align}
are satisfied. 
\end{Proposition}
\newpage
\begin{proof}
See Appendix \ref{adx: the termination condition by the contractibility of the Bellman operators}.
\end{proof}

$T_{\max}$ is always constant for every E step, $T_{\max}=\lceil(\log(1-\gamma)\varepsilon)/\log\gamma-1\rceil$. 
Thus, even if the approximation errors of $F(x,\bs{z};\bs{\theta}_{k})$ and $V(x,\bs{z};\bs{\theta}_{k})$ are smaller than $\ve$ when $t\ll T_{\max}$, 
the forward--backward algorithm cannot be terminated until $t=T_{\max}$ 
because the approximation errors of $F(x,\bs{z};\bs{\theta}_{k})$ and $V(x,\bs{z};\bs{\theta}_{k})$ cannot be evaluated in the forward--backward algorithm. 

$L_{\max}$ is adaptively determined depending on $A_{\bs{\theta}_{k}}^{L}f(x,\bs{z})$ and $B_{\bs{\theta}_{k}}^{L}v(x,\bs{z})$. 
Thus, if $A_{\bs{\theta}_{k}}^{L}f(x,\bs{z})$ and $B_{\bs{\theta}_{k}}^{L}v(x,\bs{z})$ are close enough to $F(x,\bs{z};\bs{\theta}_{k})$ and $V(x,\bs{z};\bs{\theta}_{k})$, 
the E step of MBEM can be terminated 
because the approximation errors of $F(x,\bs{z};\bs{\theta}_{k})$ and $V(x,\bs{z};\bs{\theta}_{k})$ can be evaluated owing to the contractibility of the forward and backward Bellman operators. 

Indeed, when $f(x,\bs{z})=p_{0}(x,\bs{z};\bs{\nu}_{k})$ and $v(x,\bs{z})=\bar{r}(x,\bs{z};\bs{\pi}_{k})$, 
MBEM is more efficient than EM as follows: 

\begin{Proposition}\label{prop: L is smaller than T by the contractibility of the Bellman operators}
\rm
\textls[-18]{When $f(x,\bs{z})=p_{0}(x,\bs{z};\bs{\nu}_{k})$ and $v(x,\bs{z})=\bar{r}(x,\bs{z};\bs{\pi}_{k})$, 
$L_{\max}\leq T_{\max}$ is~satisfied. }
\end{Proposition}

\begin{proof}
See Appendix \ref{adx: L is smaller than T by the contractibility of the Bellman operators}.
\end{proof}

The second property of MBEM is the arbitrariness of the initial functions, $f(x,\bs{z})$ and $v(x,\bs{z})$. 
In MBEM, the initial functions, $f(x,\bs{z})$ and $v(x,\bs{z})$, converge to the fixed points, $F(x,\bs{z};\bs{\theta}_{k})$ and $V(x,\bs{z};\bs{\theta}_{k})$, 
by applying the forward and backward Bellman operators, $A_{\bs{\theta}_{k}}$ and $B_{\bs{\theta}_{k}}$, $L_{\max}$ times. 
Therefore, if the initial functions, $f(x,\bs{z})$ and $v(x,\bs{z})$, are close to the fixed points, $F(x,\bs{z};\bs{\theta}_{k})$ and $V(x,\bs{z};\bs{\theta}_{k})$, 
$L_{\max}$ can be reduced. 
Then, the problem is what kind of the initial functions are close to the fixed points. 

We suggest that $F(x,\bs{z};\bs{\theta}_{k-1})$ and $V(x,\bs{z};\bs{\theta}_{k-1})$ are set as the initial functions, $f(x,\bs{z})$ and $v(x,\bs{z})$. 
In most cases, $\bs{\theta}_{k-1}$ is close to $\bs{\theta}_{k}$. 
When $\bs{\theta}_{k-1}$ is close to $\bs{\theta}_{k}$, 
it is expected that $F(x,\bs{z};\bs{\theta}_{k-1})$ and $V(x,\bs{z};\bs{\theta}_{k-1})$ are close to $F(x,\bs{z};\bs{\theta}_{k})$ and $V(x,\bs{z};\bs{\theta}_{k})$. 
Therefore, by setting $F(x,\bs{z};\bs{\theta}_{k-1})$ and $V(x,\bs{z};\bs{\theta}_{k-1})$ as the initial functions $f(x,\bs{z})$ and $v(x,\bs{z})$, respectively, 
$L_{\max}$ is expected to be reduced. 
Hence, MBEM can be more efficient than EM 
because MBEM can utilize the results of the previous iteration, $F(x,\bs{z};\bs{\theta}_{k-1})$ and $V(x,\bs{z};\bs{\theta}_{k-1})$, by this arbitrariness of the initial functions. 

However, it is unclear how small $L_{\max}$ can be compared to $T_{\max}$ by setting $F(x,\bs{z};\bs{\theta}_{k-1})$ and $V(x,\bs{z};\bs{\theta}_{k-1})$ as the initial functions $f(x,\bs{z})$ and $v(x,\bs{z})$. 
Therefore, numerical evaluations are needed. 
Moreover, in the first iteration, we cannot use the results of the previous iteration, $F(x,\bs{z};\bs{\theta}_{k-1})$ and $V(x,\bs{z};\bs{\theta}_{k-1})$. 
Therefore, in the first iteration, we set $f(x,\bs{z})=p_{0}(x,\bs{z};\bs{\nu}_{k})$ and $v(x,\bs{z})=\bar{r}(x,\bs{z};\bs{\pi}_{k})$ 
because these initial functions guarantee $L_{\max}\leq T_{\max}$ from Proposition \ref{prop: L is smaller than T by the contractibility of the Bellman operators}.

MBEM is summarized as Algorithm \ref{alg: MBEM}. 
The M step of Algorithm \ref{alg: MBEM} is exactly the same as that of Algorithms \ref{alg: EM} and \ref{alg: BEM}, and only the E step is different. 
The time complexity of the E step in MBEM is $\mcal{O}((|\mcal{X}||\mcal{Z}|)^2L_{\max})$. 
MBEM does not use the inverse matrix, which resolves the drawback of BEM. 
Moreover, MBEM can reduce $L_{\max}$ by the contractibility of the Bellman operators and the arbitrariness of the initial functions, which can resolve the drawback of~EM. 
\begin{algorithm}[H]
\caption{Modified Bellman EM algorithm (MBEM)}
\label{alg: MBEM}
\begin{algorithmic}
{\small
\STATE $k\leftarrow0$, Initialize $\bs{\theta}_{k}$. 
\STATE $F(x,\bs{z};\bs{\theta}_{k-1})\leftarrow p_{0}(x,\bs{z};\bs{\nu}_{k})$
\STATE $V(x,\bs{z};\bs{\theta}_{k-1})\leftarrow\bar{r}(x,\bs{z};\bs{\pi}_{k})$
\WHILE{$\bs{\theta}_{k}$ or $J(\bs{\theta}_{k})$ do not converge}
\STATE Calculate $p(x',\bs{z}'|x,\bs{z};\bs{\theta_{k}})$ by Equation (\ref{eq: p(x,z|x,z)}).
\STATE //---E step---//
\STATE $F_{0}(x,\bs{z})\leftarrow F(x,\bs{z};\bs{\theta}_{k-1})$
\STATE $V_{0}(x,\bs{z})\leftarrow F(x,\bs{z};\bs{\theta}_{k-1})$
\STATE $L\leftarrow0$
\REPEAT
\STATE $F_{L+1}(x,\bs{z})\leftarrow A_{\bs{\theta}_{k}}F_{L}(x,\bs{z})$ 
\STATE $V_{L+1}(x,\bs{z})\leftarrow B_{\bs{\theta}_{k}}V_{L}(x,\bs{z})$
\STATE $L\leftarrow L+1$
\UNTIL{$\max\{\|F_{L}-F_{L-1}\|_{1},\|V_{L}-V_{L-1}\|_{\infty}\}<\frac{1-\gamma}{\gamma}\ve$}
\STATE //---M step---//
\STATE Update $\bs{\theta}_{k}$ to $\bs{\theta}_{k+1}$ by Equations (\ref{Eq: M step-pi})--(\ref{Eq: M step-nu}).
\STATE $k\leftarrow k+1$
\ENDWHILE}
\RETURN $\bs{\theta}_{k}$
\end{algorithmic}
\end{algorithm}


\section{Summary of EM, BEM, and MBEM}\label{section: Summary of EM, BEM, and MBEM}
EM, BEM, and MBEM are summarized as in Table \ref{table: Summary of EM, BEM, and MBEM}. 
The M step is exactly the same among these algorithms, and only the E step is different:
\begin{itemize}
\item
EM obtains $F(x,\bs{z};\bs{\theta}_{k})$ and $V(x,\bs{z};\bs{\theta}_{k})$ by calculating the forward--backward algorithm up to $T_{\max}$. 
$T_{\max}$ needs to be large enough to reduce the approximation errors of $F(x,\bs{z};\bs{\theta}_{k})$ and $V(x,\bs{z};\bs{\theta}_{k})$, which impairs the computational efficiency. 
\item
BEM obtains $F(x,\bs{z};\bs{\theta}_{k})$ and $V(x,\bs{z};\bs{\theta}_{k})$ by solving the forward and backward Bellman equations. 
BEM can be more efficient than EM because BEM calculates the forward and backward Bellman equations 
instead of the forward--backward algorithm up to $T_{\max}$. 
However, BEM cannot always be more efficient than EM when the size of the state $|\mcal{X}|$ or that of the memory $|\mcal{Z}|$ is large because BEM calculates an inverse matrix to solve the forward and backward Bellman equations. 
\item
MBEM obtains $F(x,\bs{z};\bs{\theta}_{k})$ and $V(x,\bs{z};\bs{\theta}_{k})$ by applying the forward and backward Bellman operators, $A_{\bs{\theta}_{k}}$ and $B_{\bs{\theta}_{k}}$, to the initial functions, $f(x,\bs{z})$ and $v(x,\bs{z})$, $L_{\max}$ times. 
Since MBEM does not need to calculate the inverse matrix, MBEM may be more efficient than EM even when the size of problems is large, which resolves the drawback of BEM. 
Although $L_{\max}$ needs to be large enough to reduce the approximation errors of $F(x,\bs{z};\bs{\theta}_{k})$ and $V(x,\bs{z};\bs{\theta}_{k})$, which is the same problem as EM, 
MBEM can evaluate the approximation errors more tightly owing to the contractibility of $A_{\bs{\theta}_{k}}$ and $B_{\bs{\theta}_{k}}$, 
and can utilize the results of the previous iteration, $F(x,\bs{z};\bs{\theta}_{k-1})$ and $V(x,\bs{z};\bs{\theta}_{k-1})$, as the initial functions, $f(x,\bs{z})$ and $v(x,\bs{z})$. 
These properties enable MBEM to be more efficient than EM. 
\end{itemize}
\vspace{-6pt}

\end{paracol}
    \begin{specialtable}[H]
    \tablesize{\footnotesize}
    \widetable
\caption{Summary of EM, BEM, and MBEM. }
\label{table: Summary of EM, BEM, and MBEM}
    \setlength{\cellWidtha}{\columnwidth/4-2\tabcolsep-.8in}
\setlength{\cellWidthb}{\columnwidth/4-2\tabcolsep+0.0in}
\setlength{\cellWidthc}{\columnwidth/4-2\tabcolsep+.4in}
\setlength{\cellWidthd}{\columnwidth/4-2\tabcolsep+.4in}
\scalebox{1}[1]{\begin{tabularx}{\columnwidth}{>{\PreserveBackslash\centering}m{\cellWidtha}>{\PreserveBackslash\centering}m{\cellWidthb}>{\PreserveBackslash\centering}m{\cellWidthc}>{\PreserveBackslash\centering}m{\cellWidthd}}
\toprule
	&\textbf{EM}&\textbf{Bellman EM (BEM)}&\textbf{Modified Bellman EM (MBEM)}\\ \midrule
	E step
	&forward--backward algorithm\linebreak
		$F(x,\bs{z};\bs{\theta}_{k})\approx\sum_{t=0}^{T_{\max}}\gamma^{t}\alpha_{t}(x,\bs{z};\bs{\theta}_{k})$\linebreak
		$V(x,\bs{z};\bs{\theta}_{k})\approx\sum_{t=0}^{T_{\max}}\gamma^{t}\beta_{t}(x,\bs{z};\bs{\theta}_{k})$\linebreak
		$\mcal{O}((|\mcal{X}||\mcal{Z}|)^2T_{\max})$
	&forward and backward Bellman equations\linebreak
		$\bs{F}(\bs{\theta}_{k})=(\bs{I}-\gamma\bs{P}(\bs{\theta}_{k}))^{-1}\bs{p}(\bs{\nu}_{k})$\linebreak
		$\bs{V}(\bs{\theta}_{k})=((\bs{I}-\gamma\bs{P}(\bs{\theta}_{k}))^{-1})^{T}\bs{r}(\bs{\pi}_{k})$\linebreak
		$\mcal{O}((|\mcal{X}||\mcal{Z}|)^3)$ 
	&forward and backward Bellman operators\linebreak 
		$F(x,\bs{z};\bs{\theta}_{k})\approx A_{\bs{\theta}_{k}}^{L_{\max}}f(x,\bs{z})$\linebreak
		$V(x,\bs{z};\bs{\theta}_{k})\approx B_{\bs{\theta}_{k}}^{L_{\max}}v(x,\bs{z})$\linebreak
		$\mcal{O}((|\mcal{X}||\mcal{Z}|)^2L_{\max})$\\ \midrule
	M step
	&Equations (\ref{Eq: M step-pi})--(\ref{Eq: M step-nu})\linebreak
		$\mcal{O}((|\mcal{X}||\mcal{Z}|)^2|\mcal{Y}||\mcal{A}|)$ 
	&Equations (\ref{Eq: M step-pi})--(\ref{Eq: M step-nu})\linebreak
		$\mcal{O}((|\mcal{X}||\mcal{Z}|)^2|\mcal{Y}||\mcal{A}|)$ 
	&Equations (\ref{Eq: M step-pi})--(\ref{Eq: M step-nu})\linebreak
		$\mcal{O}((|\mcal{X}||\mcal{Z}|)^2|\mcal{Y}||\mcal{A}|)$\\
		\bottomrule
\end{tabularx}
}
\end{specialtable}
\begin{paracol}{2}
\switchcolumn

\section{Numerical Experiment}\label{section: Numerical experiment}

In this section, we compare the performance of EM, BEM, and MBEM using numerical experiments of four benchmarks for DEC-POMDP: 
broadcast \cite{hansen2004dynamic}, recycling robot \cite{amato2012optimizing}, wireless network \cite{pajarinen2011a}, and box pushing \cite{seuken2012improved}. 
Detailed settings such as the state transition probability, the observation probability, and the reward function are described at \url{http://masplan.org/problem\_domains}, accessed on 22nd June, 2020. 
We implement EM, BEM, and MBEM in~C++. 

Figure \ref{Fig: experiment} shows the experimental results. 
In all the experiments, we set the number of agent $N=2$, the discount factor $\gamma=0.99$, the upper bound of the approximation error $\varepsilon=0.1$, and the size of the memory available to the $i$th agent $|\mcal{Z}^{i}|=2$. 
The size of the state $|\mcal{X}|$, the action $|\mcal{A}^{i}|$, and the observation $|\mcal{Y}^{i}|$ are different for each problem, which are shown on each panel. 
We note that the size of the state $|\mcal{X}|$ is small in the broadcast (a,e,i) and the recycling robot (b,f,j), whereas it is large in the wireless network (c,g,k) and the box pushing (d,h,l). 
\end{paracol}
\begin{figure}[H]
\widefigure
    (\textbf{a})\hspace{-3.0mm}
	\begin{minipage}[t][][b]{40mm}
		\includegraphics[width=40mm,height=30mm]{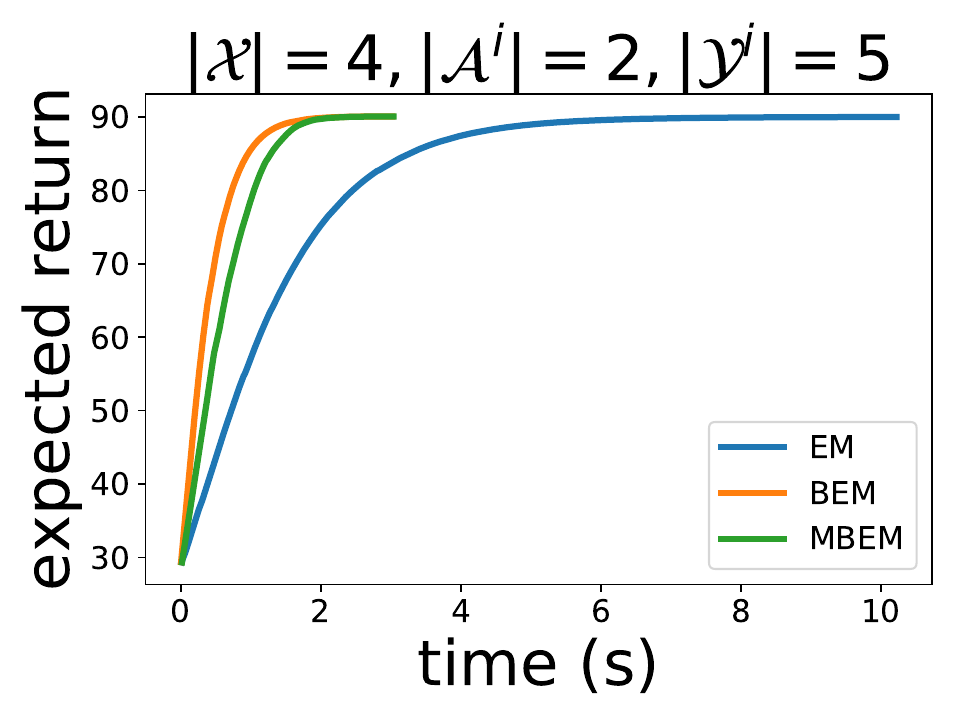}
	\end{minipage}
	(\textbf{b})\hspace{-3.0mm}
	\begin{minipage}[t][][b]{40mm}
		\includegraphics[width=40mm,height=30mm]{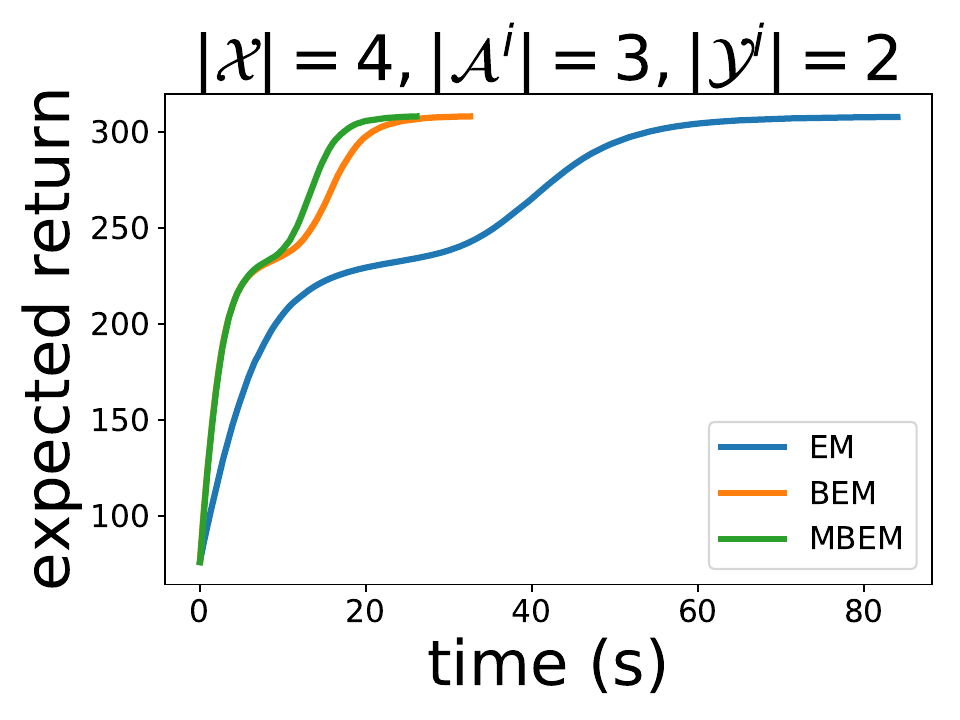}
	\end{minipage}
	(\textbf{c})\hspace{-3.0mm}
	\begin{minipage}[t][][b]{40mm}
		\includegraphics[width=40mm,height=30mm]{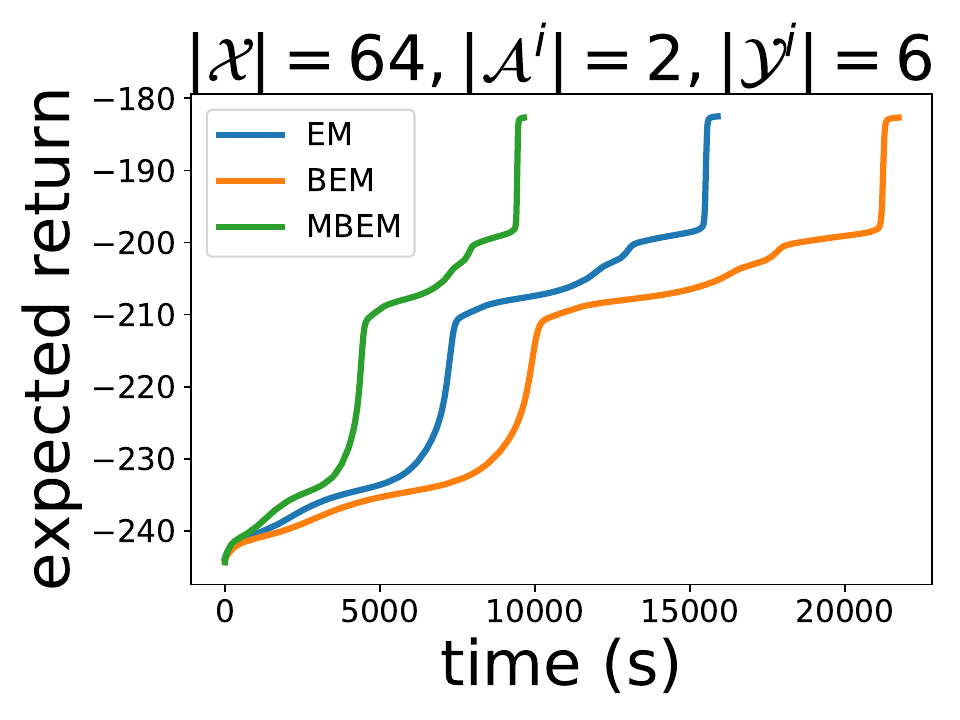}
	\end{minipage}
	(\textbf{d})\hspace{-3.0mm}
	\begin{minipage}[t][][b]{40mm}
		\includegraphics[width=40mm,height=30mm]{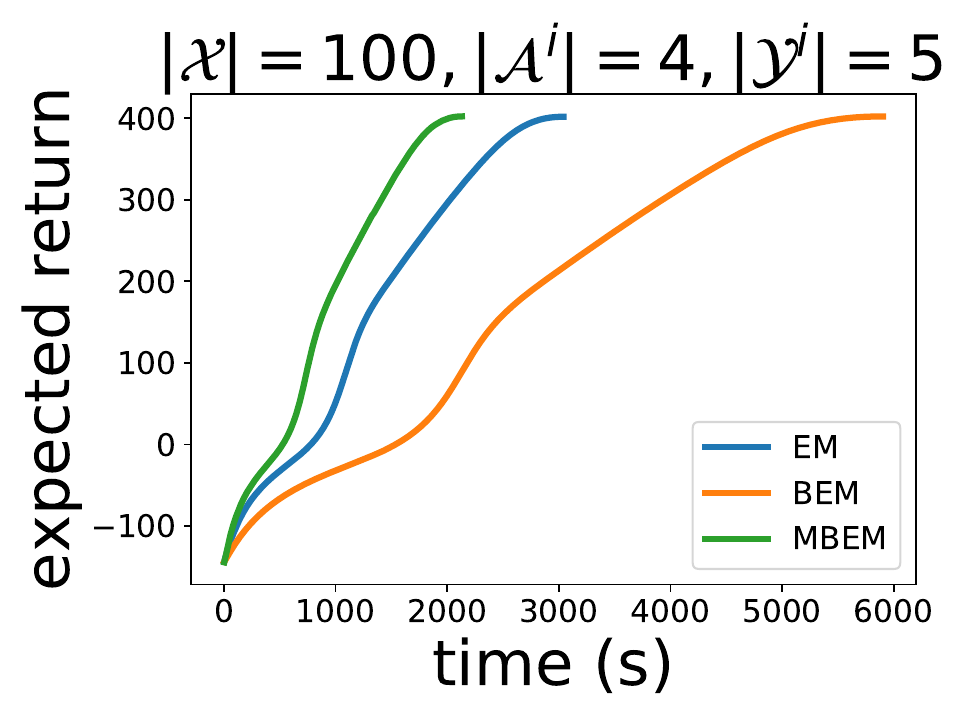}
	\end{minipage}\\
	(\textbf{e})\hspace{-3.0mm}
	\begin{minipage}[t][][b]{40mm}
		\includegraphics[width=40mm,height=30mm]{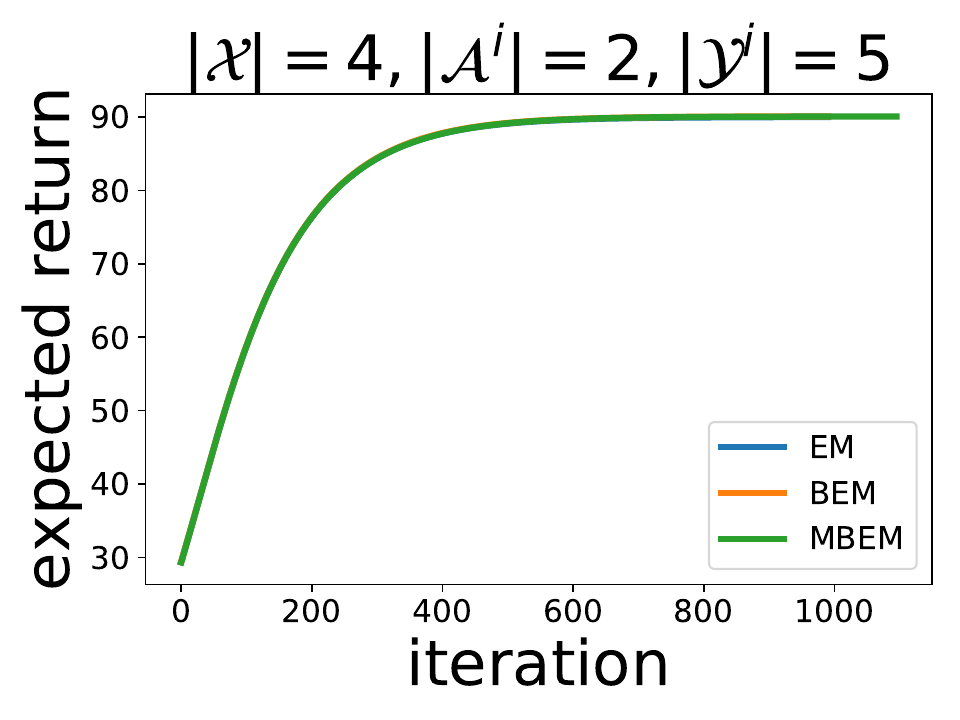}
	\end{minipage}
	(\textbf{f})\hspace{-3.0mm}
	\begin{minipage}[t][][b]{40mm}
		\includegraphics[width=40mm,height=30mm]{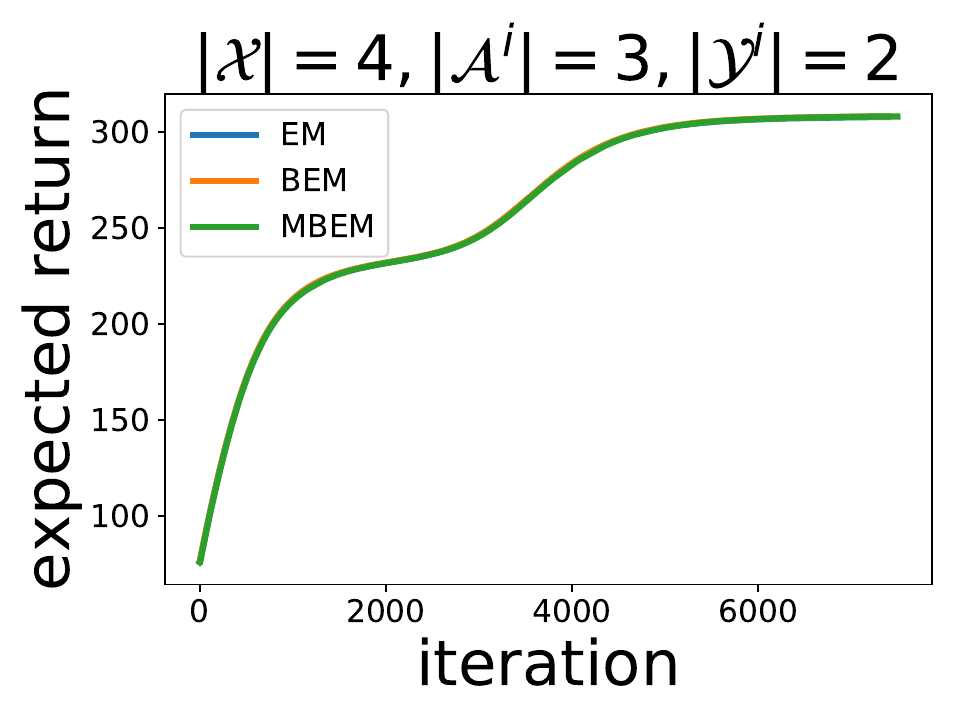}
	\end{minipage}
	(\textbf{g})\hspace{-3.0mm}
	\begin{minipage}[t][][b]{40mm}
		\includegraphics[width=40mm,height=30mm]{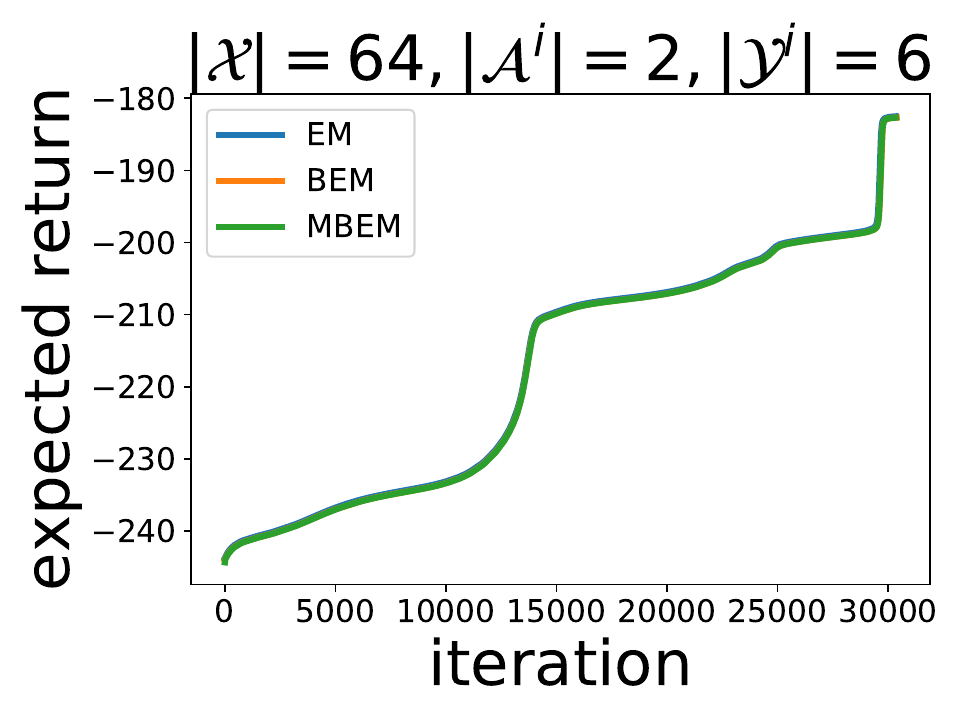}
	\end{minipage}
	(\textbf{h})\hspace{-3.0mm}
	\begin{minipage}[t][][b]{40mm}
		\includegraphics[width=40mm,height=30mm]{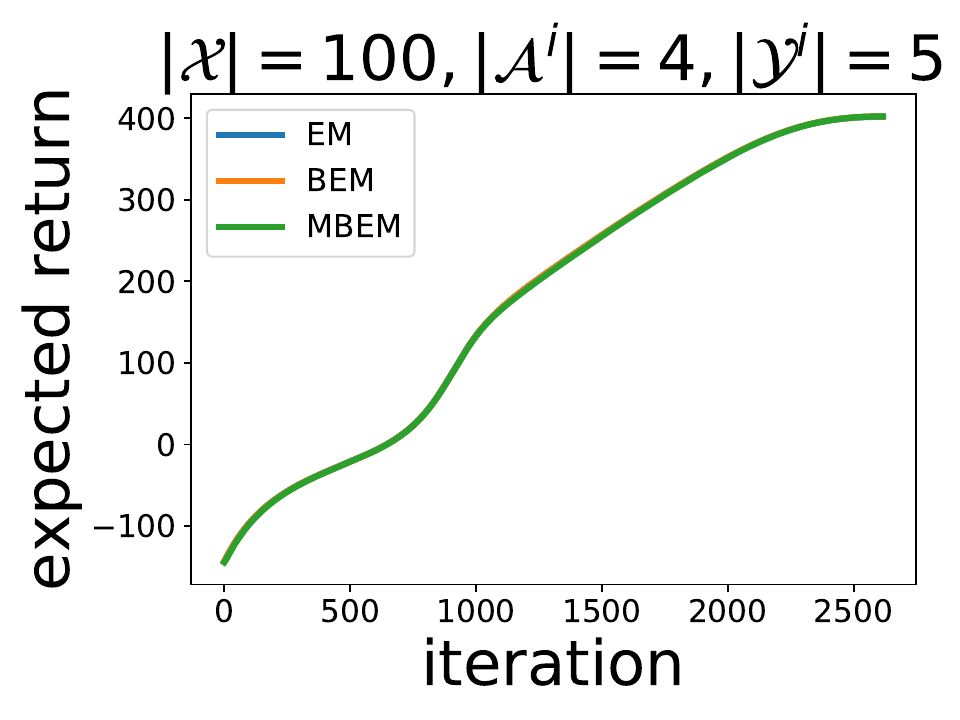}
	\end{minipage}\\
	(\textbf{i})\hspace{-3.0mm}
	\begin{minipage}[t][][b]{40mm}
		\includegraphics[width=40mm,height=30mm]{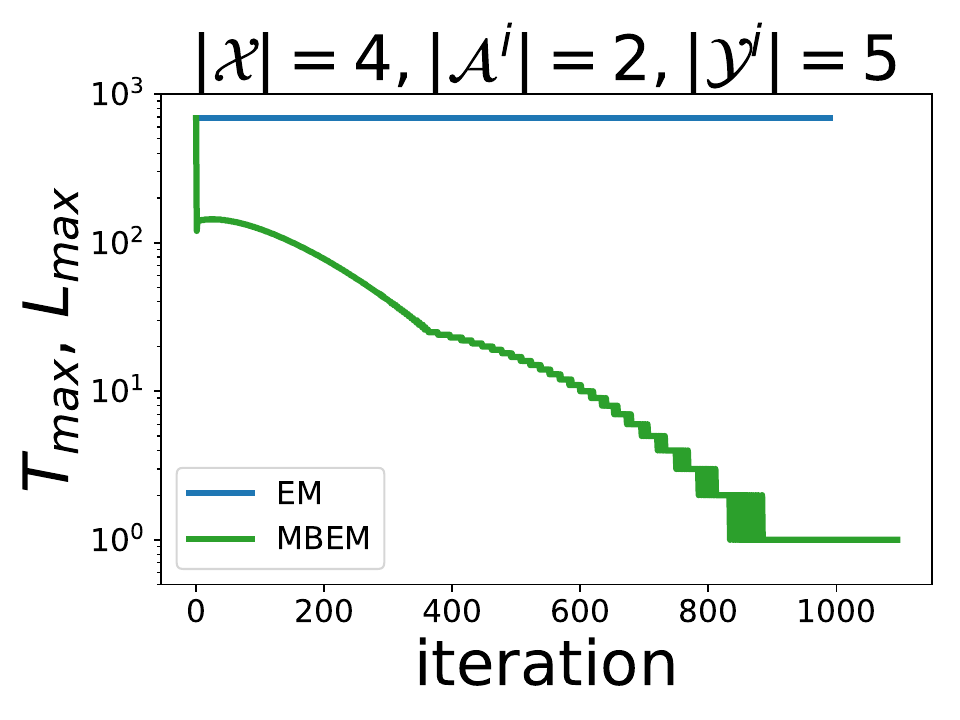}
	\end{minipage}
	(\textbf{j})\hspace{-3.0mm}
	\begin{minipage}[t][][b]{40mm}
		\includegraphics[width=40mm,height=30mm]{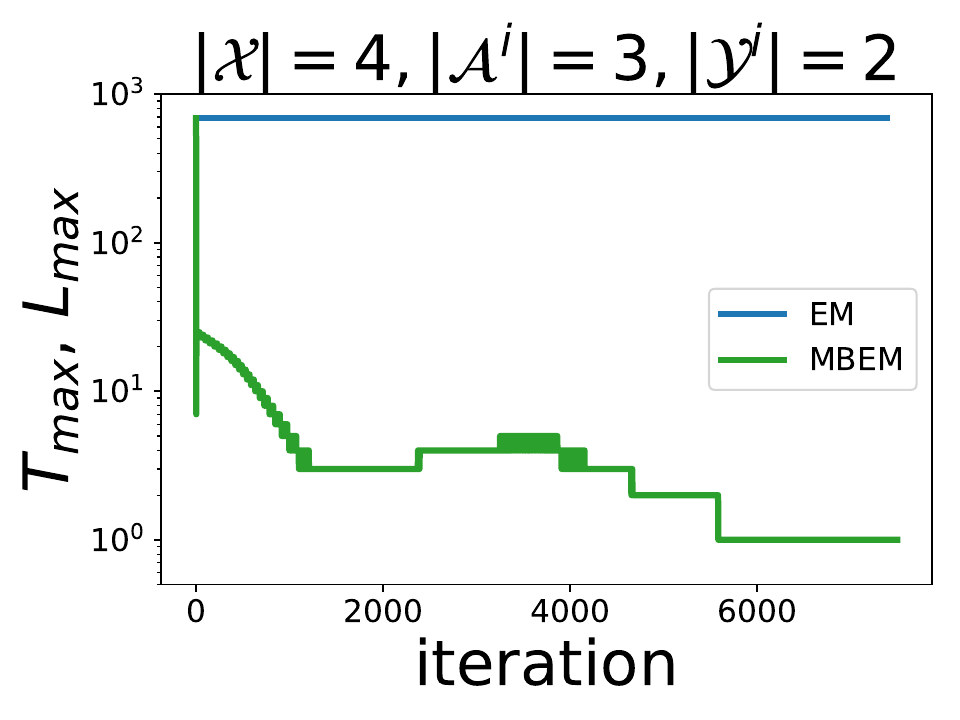}
	\end{minipage}
	(\textbf{k})\hspace{-3.0mm}
	\begin{minipage}[t][][b]{40mm}
		\includegraphics[width=40mm,height=30mm]{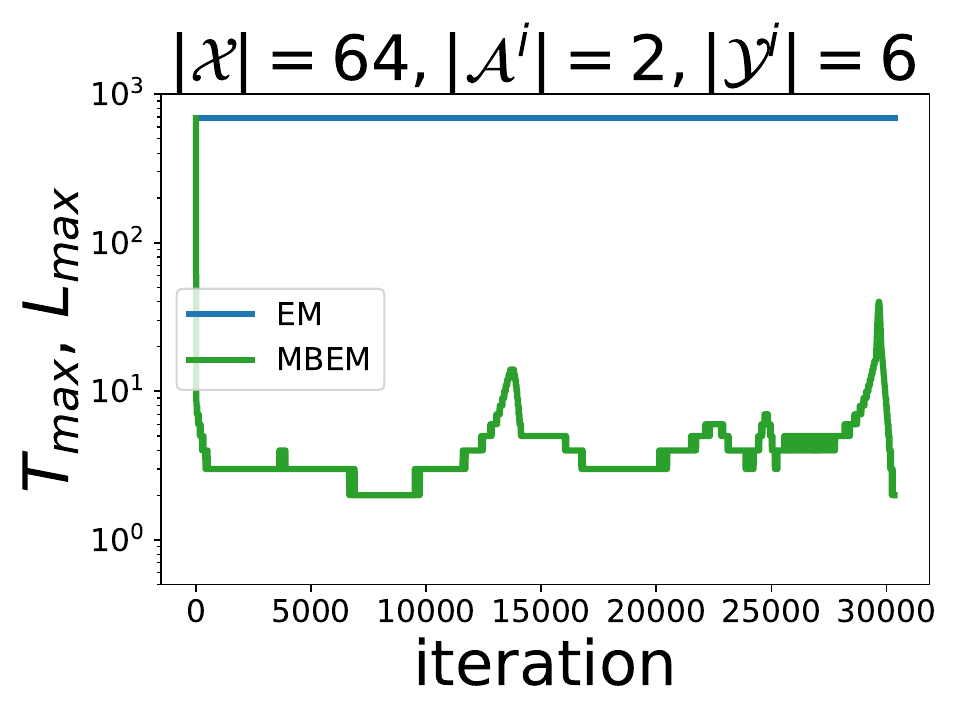}
	\end{minipage}
	(\textbf{l})\hspace{-3.0mm}
	\begin{minipage}[t][][b]{40mm}
		\includegraphics[width=40mm,height=30mm]{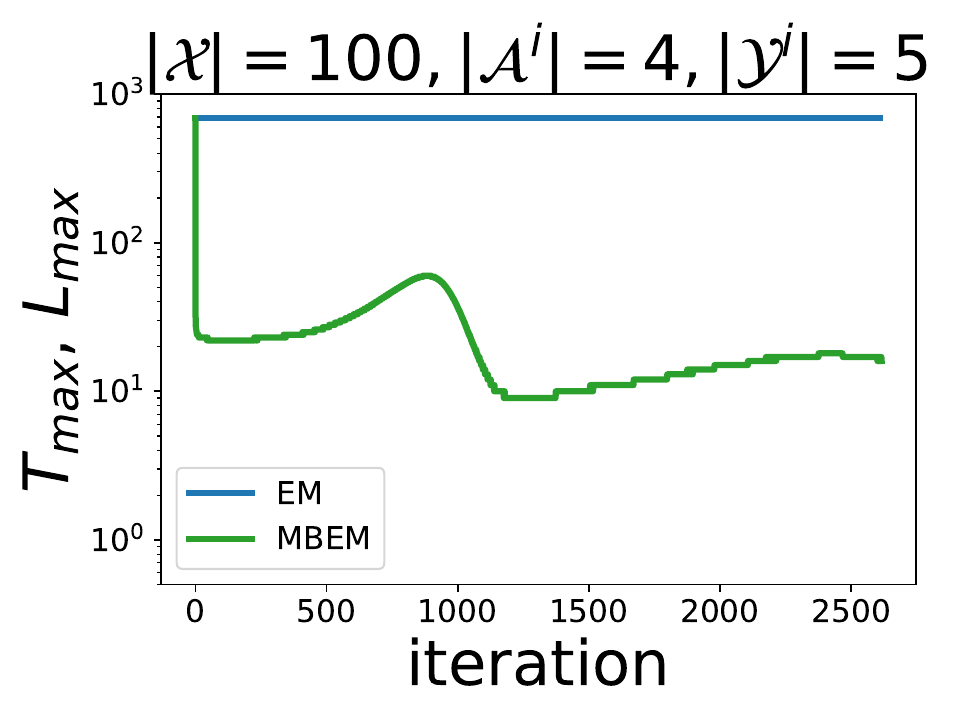}
	\end{minipage}\\
	\caption{
	Experimental results of four benchmarks for DEC-POMDP: 
	(\textbf{a},\textbf{e},\textbf{i}) broadcast; 
	(\textbf{b},\textbf{f},\textbf{j}) recycling robot; 
	(\textbf{c},\textbf{g},\textbf{k})~wireless network; 
	(\textbf{d},\textbf{h},\textbf{l}) box pushing. 
	(\textbf{a}--\textbf{d}) The expected return $J(\bs{\theta}_{k})$ as a function of the computational time. 
	(\textbf{e}--\textbf{h})~The expected return $J(\bs{\theta}_{k})$ as a function of the iteration $k$. 
	(\textbf{i}--\textbf{l}) $T_{\max}$ and $L_{\max}$ as functions of the iteration $k$. 
	In all the experiments, we set the number of agent $N=2$, the discount factor $\gamma=0.99$, the upper bound of the approximation error $\varepsilon=0.1$, 
	and the size of the memory available to the $i$th agent $|\mcal{Z}^{i}|=2$. 
	The size of the state $|\mcal{X}|$, the action $|\mcal{A}^{i}|$, and the observation $|\mcal{Y}^{i}|$ are different for each problem, which are shown on each~panel. 
	}
	\label{Fig: experiment}
\end{figure}

\begin{paracol}{2}
\switchcolumn

While the expected return $J(\bs{\theta}_{k})$ with respect to the computational time is different between the algorithms (a--d), that with respect to the iteration $k$ is almost the same (e--h). 
This is because the M step of these algorithms is exactly the same. 
Therefore, the difference of the computational time is caused by the computational time of the E step. 

The convergence of BEM is faster than that of EM in the small state size problems, i.e.,
 Figure \ref{Fig: experiment}a,b. 
This is because EM calculates the forward--backward algorithm from $t=0$ to $t=T_{\max}$, where $T_{\max}$ is large. 
On the other hand, the convergence of BEM is slower than that of EM in the large state size problems, {i.e.}, Figure \ref{Fig: experiment}c,d. 
This is because BEM calculates the inverse matrix.  

\textls[+12]{{The} convergence of MBEM is faster than that of EM in all the experiments in\linebreak Figure \ref{Fig: experiment}a--d. }
This is because $L_{\max}$ is smaller than $T_{\max}$ as shown in Figure \ref{Fig: experiment}i--l. 
While EM requires about 1000 calculations of the forward--backward algorithm to guarantee that the approximation error of $F(x,\bs{z};\bs{\theta}_{k})$ and $V(x,\bs{z};\bs{\theta}_{k})$ is smaller than $\ve$, 
MBEM requires only about 10 calculations of the forward and backward Bellman operators. 
Thus, MBEM is more efficient than EM. 
The reason why $L_{\max}$ is smaller than $T_{\max}$ is that MBEM can utilize the results of the previous iteration, $F(x,\bs{z};\bs{\theta}_{k-1})$ and $V(x,\bs{z};\bs{\theta}_{k-1})$, 
as the initial functions, $f(x,\bs{z})$ and $v(x,\bs{z})$. 
It is shown from $L_{\max}$ and $T_{\max}$ in the first iteration. 
In the first iteration $k=0$, $L_{\max}$ is almost the same with $T_{\max}$ 
because $F(x,\bs{z};\bs{\theta}_{k-1})$ and $V(x,\bs{z};\bs{\theta}_{k-1})$ cannot be used as the initial functions $f(x,\bs{z})$ and $v(x,\bs{z})$ in the first iteration. 
On the other hand, in the subsequent iterations $k\geq1$, $L_{\max}$ is much smaller than $T_{\max}$ 
because MBEM can utilize the results of the previous iteration, $F(x,\bs{z};\bs{\theta}_{k-1})$ and $V(x,\bs{z};\bs{\theta}_{k-1})$, the initial functions $f(x,\bs{z})$ and $v(x,\bs{z})$.

\section{Conclusions and Future works}\label{section: Conclusion}

In this paper, we propose the Bellman EM algorithm (BEM) and the modified Bellman EM algorithm (MBEM) 
by introducing the forward and backward Bellman equations into the EM algorithm for DEC-POMDP. 
BEM can be more efficient than EM because BEM does not calculate the forward--backward algorithm up to the infinite horizon. 
However, BEM cannot always be more efficient than EM when the size of the state or that of the memory is large because BEM calculates the inverse matrix. 
MBEM can be more efficient than EM regardless of the size of problems because MBEM does not calculate the inverse matrix. 
Although MBEM needs to calculate the forward and backward Bellman operators infinite times, 
MBEM can evaluate the approximation errors more tightly owing to the contractibility of these operators, 
and can utilize the results of the previous iteration owing to the arbitrariness of initial functions, which enables MBEM to be more efficient than EM. 
We verified this theoretical evaluation by the numerical experiment, which demonstrates that the convergence of MBEM is much faster than that of EM regardless of the size of~problems.

Our algorithms still leave room for further improvements that deal with the real-world problems, which often have a large discrete or continuous state space. 
Some of them may be addressed by the advanced techniques of the Bellman equations \cite{bertsekas2000dynamic,puterman2014markov,sutton1998,sutton2018reinforcement}. 
For example, MBEM may be accelerated by the Gauss--Seidel method \cite{puterman2014markov}. 
The convergence rate of the E step of MBEM is given by the discount factor $\gamma$, which is the same as that of EM. 
However, the Gauss--Seidel method modifies the Bellman operators, which allows the convergence rate of MBEM to be smaller than the discount factor $\gamma$. 
Therefore, even if $F(x,\bs{z};\bs{\theta}_{k-1})$ and $V(x,\bs{z};\bs{\theta}_{k-1})$ are not close to $F(x,\bs{z};\bs{\theta}_{k})$ and $V(x,\bs{z};\bs{\theta}_{k})$, MBEM may be more efficient than EM by the Gauss--Seidel method. 
Moreover, in DEC-POMDP with a large discrete or continuous state space, $F(x,\bs{z};\bs{\theta}_{k})$ and $V(x,\bs{z};\bs{\theta}_{k})$ cannot be expressed exactly because it requires a large space complexity. 
This problem may be resolved by the value function approximation \cite{bertsekas2011approximate,liu2015feature,mnih2015human}. 
The value function approximation approximates $F(x,\bs{z};\bs{\theta}_{k})$ and $V(x,\bs{z};\bs{\theta}_{k})$ using parametric models such as neural networks. 
The problem is how to find the optimal approximate parameters.
The value function approximation finds them by the Bellman equation. 
Therefore, the potential extensions of our algorithms may lead to the applications to the real-world DEC-POMDP problems. 

\vspace{6pt} 
\authorcontributions{Conceptualization, T.T., T.J.K.; Formal analysis, T.T., T.J.K.; Funding acquisition, T.J.K.; Writing---original draft, T.T., T.J.K. All authors have read and agreed to the published version of the manuscript.} 

\funding{This research is supported by JSPS KAKENHI Grant Number 19H05799 and by JST CREST Grant Number JPMJCR2011.} 

\acknowledgments{
We would like to thank the lab members for a fruitful discussion.}

\conflictsofinterest{The authors declare no conflict of interest. }

\newpage
\appendixtitles{yes} 
\appendixstart
\appendix
\section{Proof in Section~\ref{section: EM algorithm for DEC-POMDP}}
\subsection{Proof of Theorem \ref{th: control as inference}}\label{adx: control as inference}
$J(\bs{\theta})$ can be calculated as follows: 
\begin{align}
    J(\bs{\theta})
    :=&\mb{E}_{\bs{\theta}}\left[\sum_{t=0}^{\infty}\gamma^{t}r(x_{t},\bs{a}_{t})\right]\nonumber\\
    =&\sum_{x_{0:\infty},\bs{a}_{0:\infty}}p(x_{0:\infty},\bs{a}_{0:\infty};\bs{\theta})\sum_{t=0}^{\infty}\gamma^{t}r(x_{t},\bs{a}_{t})
    =\sum_{t=0}^{\infty}\gamma^{t}\sum_{x_{t},\bs{a}_{t}}p(x_{t},\bs{a}_{t};\bs{\theta})r(x_{t},\bs{a}_{t})\nonumber\\
    =&(1-\gamma)^{-1}\sum_{T=0}^{\infty}p(T)\sum_{x_{T},\bs{a}_{T}}p(x_{T},\bs{a}_{T};\bs{\theta})
    \left[(r_{\max}-r_{\min})p(o=1|x_{T},\bs{a}_{T})+r_{\min}\right]\nonumber\\
    =&(1-\gamma)^{-1}\left[(r_{\max}-r_{\min})p(o=1;\bs{\theta})+r_{\min}\right]
\end{align}
where $x_{0:\infty}:=\{x_{0},...,x_{\infty}\}$ and $\bs{a}_{0:\infty}:=\{\bs{a}_{0},...,\bs{a}_{\infty}\}$. 

\subsection{Proof of Proposition \ref{th: M step}}\label{adx: M step}
In order to prove Proposition \ref{th: M step}, we calculate $Q(\bs{\theta};\bs{\theta}_{k})$. It can be calculated as follows: \begin{align}
	Q(\bs{\theta};\bs{\theta}_{k})
	:=&\mb{E}_{\bs{\theta}_{k}}\left[\left.\log p(o=1,x_{0:T},\bs{y}_{0:T},\bs{z}_{0:T},\bs{a}_{0:T},T;\bs{\theta})\right|o=1\right]\nonumber\\
	=&\mb{E}_{\bs{\theta}_{k}}\left[\left.\sum_{i=1}^{N}\left\{\sum_{t=0}^{T}\log \pi^{i}(a_{t}^{i}|z_{t}^{i})+\sum_{t=1}^{T}\log \lambda^{i}(z_{t}^{i}|z_{t-1}^{i},y_{t}^{i})+\log \nu^{i}(z_{0}^{i})\right\}\right|o=1\right]+C\nonumber\\
	=&\sum_{i=1}^{N}\left(Q(\pi^{i};\bs{\theta}_{k})+Q(\lambda^{i};\bs{\theta}_{k})+Q(\nu^{i};\bs{\theta}_{k})\right)+C, 
\end{align}
where
\begin{align}
	Q(\pi^{i};\bs{\theta}_{k})&:=\mb{E}_{\bs{\theta}_{k}}\left[\left.\sum_{t=0}^{T}\log \pi^{i}(a_{t}^{i}|z_{t}^{i})\right|o=1\right],\\
	Q(\lambda^{i};\bs{\theta}_{k})&:=\mb{E}_{\bs{\theta}_{k}}\left[\left.\sum_{t=1}^{T}\log \lambda^{i}(z_{t}^{i}|z_{t-1}^{i},y_{t}^{i})\right|o=1\right],\\
	Q(\nu^{i};\bs{\theta}_{k})&:=\mb{E}_{\bs{\theta}_{k}}\left[\left.\log \nu^{i}(z_{0}^{i})\right|o=1\right].
\end{align}
$C$ is a constant independent of $\bs{\theta}$. 

\begin{Proposition}[\cite{kumar2010}]
\label{prop: Q-function}
\rm
$Q(\pi^{i};\bs{\theta}_{k})$, $Q(\lambda^{i};\bs{\theta}_{k})$, and $Q(\nu^{i};\bs{\theta}_{k})$ are calculated as follows:
\begin{align}
	Q(\pi^{i};\bs{\theta}_{k})
	&\propto\sum_{\bs{a},\bs{z}}\bs{\pi}_{k}(\bs{a}|\bs{z})\log \pi^{i}(a^{i}|z^{i})\nonumber\\
	&\times\sum_{x,x',\bs{z}'}p(x',\bs{z}'|x,\bs{z},\bs{a};\bs{\lambda}_{k})
	F(x,\bs{z};\bs{\theta}_{k})(\bar{r}(x,\bs{a})+\gamma V(x',\bs{z}';\bs{\theta}_{k}))\label{eq: Q(pi)},\\
	Q(\lambda^{i};\bs{\theta}_{k})
	&\propto\sum_{\bs{z}',\bs{z},\bs{y}'}\bs{\lambda}_{k}(\bs{z}'|\bs{z},\bs{y}')\log \lambda^{i}(z^{i'}|z^{i},y^{i'})\nonumber\\
	&\times\sum_{x',x}p(x',\bs{y}'|x,\bs{z};\bs{\pi}_{k})
	F(x,\bs{z};\bs{\theta}_{k})V(x',\bs{z}';\bs{\theta}_{k})\label{eq: Q(lambda)},\\
	Q(\nu^{i};\bs{\theta}_{k})
	&\propto\sum_{\bs{z}}p_{0}(\bs{z};\bs{\nu}_{n})\log \nu^{i}(z^{i})
	\sum_{x}p_{0}(x)V(x,\bs{z};\bs{\theta}_{k})\label{eq: Q(nu)}.
\end{align}
$F(x,\bs{z};\bs{\theta}_{k})$ and $V(x,\bs{z};\bs{\theta}_{k})$ are defined by Equations (\ref{eq: forward var}) and (\ref{eq: backward var}). 
\end{Proposition}

\begin{proof}
Firstly, we prove Equation (\ref{eq: Q(pi)}). $Q(\pi^{i};\bs{\theta}_{k})$ can be calculated as follows:
\begin{align}
	&Q(\pi^{i};\bs{\theta}_{k})
	:=\mb{E}_{\bs{\theta}_{k}}\left[\left.\sum_{t=0}^{T}\log \pi^{i}(a_{t}^{i}|z_{t}^{i})\right|o=1\right]\nonumber\\
	&=\frac{1}{p(o=1;\bs{\theta}_{k})}\sum_{T=0}^{\infty}\sum_{x_{0:T},\bs{y}_{0:T},\bs{z}_{0:T},\bs{a}_{0:T}}p(o=1,x_{0:T},\bs{y}_{0:T},\bs{z}_{0:T},\bs{a}_{0:T},T;\bs{\theta}_{k})\sum_{t=0}^{T}\log \pi^{i}(a_{t}^{i}|z_{t}^{i})\nonumber\\
	&=\frac{1}{p(o=1;\bs{\theta}_{k})}\sum_{T=0}^{\infty}\sum_{t=0}^{T}\sum_{x_{t},\bs{z}_{t},\bs{a}_{t}}p(o=1,x_{t},\bs{z}_{t},\bs{a}_{t},T;\bs{\theta}_{k})\log \pi^{i}(a_{t}^{i}|z_{t}^{i})\nonumber\\
	&=\frac{1}{p(o=1;\bs{\theta}_{k})}\sum_{T=0}^{\infty}\sum_{t=0}^{T}\sum_{x_{t},\bs{z}_{t},\bs{a}_{t}}p(T)p_{t}(x_{t},\bs{z}_{t};\bs{\theta}_{k})\bs{\pi}_{k}(\bs{a}_{t}|\bs{z}_{t})\nonumber\\
	&\times p_{t}(o=1|x_{t},\bs{z}_{t},\bs{a}_{t},T;\bs{\theta}_{k})\log \pi^{i}(a_{t}^{i}|z_{t}^{i})\nonumber\\
	&=\frac{1}{p(o=1;\bs{\theta}_{k})}\sum_{T=0}^{\infty}\sum_{t=0}^{T}\sum_{x,\bs{z},\bs{a}}p(T)p_{t}(x,\bs{z};\bs{\theta}_{k})\bs{\pi}_{k}(\bs{a}|\bs{z})p_{t}(o=1|x,\bs{z},\bs{a},T;\bs{\theta}_{k})\log \pi^{i}(a^{i}|z^{i})\nonumber\\
	&=\frac{1}{p(o=1;\bs{\theta}_{k})}\sum_{x,\bs{z},\bs{a}}\bs{\pi}_{k}(\bs{a}|\bs{z})\log \pi^{i}(a^{i}|z^{i})\nonumber\\
	&\times \sum_{T=0}^{\infty}p(T)\sum_{t=0}^{T}p_{t}(x,\bs{z};\bs{\theta}_{k})p_{t}(o=1|x,\bs{z},\bs{a},T;\bs{\theta}_{k}).
\end{align}
Since $\sum_{T=0}^{\infty}\sum_{t=0}^{T}...=\sum_{t=0}^{T}\sum_{T=t}^{\infty}...$, we have
\begin{align}
	Q(\pi^{i};\bs{\theta}_{k})
	=&\frac{1}{p(o=1;\bs{\theta}_{k})}\sum_{x,\bs{z},\bs{a}}\bs{\pi}_{k}(\bs{a}|\bs{z})\log \pi^{i}(a^{i}|z^{i})\nonumber\\
	&\times\sum_{t=0}^{\infty}p_{t}(x,\bs{z};\bs{\theta}_{k})\sum_{T=t}^{\infty}p(T)p_{t}(o=1|x,\bs{z},\bs{a},T;\bs{\theta}_{k})\nonumber\\
	=&\frac{1-\gamma}{p(o=1;\bs{\theta}_{k})}\sum_{x,\bs{z},\bs{a}}\bs{\pi}_{k}(\bs{a}|\bs{z})\log \pi^{i}(a^{i}|z^{i})\nonumber\\
	&\times\sum_{t=0}^{\infty}p_{t}(x,\bs{z};\bs{\theta}_{k})\sum_{T=t}^{\infty}\gamma^{T}p_{t}(o=1|x,\bs{z},\bs{a},T;\bs{\theta}_{k}).
\end{align}
Since $p_{t}(o=1|x,\bs{z},\bs{a},T;\bs{\theta}_{k})=p_{0}(o=1|x,\bs{z},\bs{a},T-t;$ $\bs{\theta}_{k})$, 
\begin{align}
	Q(\pi^{i};\bs{\theta}_{k})
	=&\frac{1-\gamma}{p(o=1;\bs{\theta}_{k})}\sum_{x,\bs{z},\bs{a}}\bs{\pi}_{k}(\bs{a}|\bs{z})\log \pi^{i}(a^{i}|z^{i})\nonumber\\
	&\times\sum_{t=0}^{\infty}p_{t}(x,\bs{z};\bs{\theta}_{k})\sum_{T=t}^{\infty}\gamma^{T}p_{0}(o=1|x,\bs{z},\bs{a},T-t;\bs{\theta}_{k})\nonumber\\
	=&\frac{1-\gamma}{p(o=1;\bs{\theta}_{k})}\sum_{x,\bs{z},\bs{a}}\bs{\pi}_{k}(\bs{a}|\bs{z})\log \pi^{i}(a^{i}|z^{i})\nonumber\\
	&\times\sum_{t=0}^{\infty}\gamma^{t}p_{t}(x,\bs{z};\bs{\theta}_{k})\sum_{T=0}^{\infty}\gamma^{T}p_{0}(o=1|x,\bs{z},\bs{a},T;\bs{\theta}_{k})\nonumber\\
	=&\frac{1-\gamma}{p(o=1;\bs{\theta}_{k})}\sum_{x,\bs{z},\bs{a}}\bs{\pi}_{k}(\bs{a}|\bs{z})\log \pi^{i}(a^{i}|z^{i})\nonumber\\
	&\times F(x,\bs{z};\bs{\theta}_{k})\sum_{T=0}^{\infty}\gamma^{T}p_{0}(o=1|x,\bs{z},\bs{a},T;\bs{\theta}_{k})\label{eq: Q(pi)-tmp1}.
\end{align}
$\sum_{T=0}^{\infty}\gamma^{T}p_{0}(o=1|x,\bs{z},\bs{a},T;\bs{\theta}_{k})$ can be calculated as follows: 
\begin{align}
	&\sum_{T=0}^{\infty}\gamma^{T}p_{0}(o=1|x,\bs{z},\bs{a},T;\bs{\theta}_{k})\nonumber\\
	=&\bar{r}(x,\bs{a})+\sum_{T=1}^{\infty}\gamma^{T}p_{0}(o=1|x,\bs{z},\bs{a},T;\bs{\theta}_{k})\nonumber\\
	=&\bar{r}(x,\bs{a})+\sum_{T=1}^{\infty}\gamma^{T}\sum_{x',\bs{z}'}p_{1}(o=1|x',\bs{z}',T;\bs{\theta}_{k})p(x',\bs{z}'|x,\bs{z},\bs{a};\bs{\lambda}_{k})\nonumber\\
	=&\bar{r}(x,\bs{a})+\gamma\sum_{x',\bs{z}'}p(x',\bs{z}'|x,\bs{z},\bs{a};\bs{\lambda}_{k})\sum_{T=0}^{\infty}\gamma^{T}p_{0}(o=1|x',\bs{z}',T;\bs{\theta}_{k})\nonumber\\
	=&\bar{r}(x,\bs{a})+\gamma\sum_{x',\bs{z}'}p(x',\bs{z}'|x,\bs{z},\bs{a};\bs{\lambda}_{k})V(x',\bs{z}';\bs{\theta}_{k})\label{eq: Q(pi)-tmp2}.
\end{align}
From Equations (\ref{eq: Q(pi)-tmp1}) and (\ref{eq: Q(pi)-tmp2}), we obtain
\begin{align}
	Q(\pi^{i};\bs{\theta}_{k})
	=&\frac{1-\gamma}{p(o=1;\bs{\theta}_{k})}\sum_{\bs{a},\bs{z}}\bs{\pi}_{k}(\bs{a}|\bs{z})\log \pi^{i}(a^{i}|z^{i})\nonumber\\
	&\times\sum_{x,x',\bs{z}'}p(x',\bs{z}'|x,\bs{z},\bs{a};\bs{\lambda}_{k})
	F(x,\bs{z};\bs{\theta}_{k})
	(\bar{r}(x,\bs{a})+\gamma V(x',\bs{z}';\bs{\theta}_{k})).
\end{align}
Therefore, Equation (\ref{eq: Q(pi)}) is proved. \\

Secondly, we prove Equation (\ref{eq: Q(lambda)}). $Q(\lambda^{i};\bs{\theta}_{k})$ can be calculated as follows: 
\begin{align}
	&Q(\lambda^{i};\bs{\theta}_{k})
	:=\mb{E}_{\bs{\theta}_{k}}\left[\left.\sum_{t=1}^{T}\log \lambda^{i}(z_{t}^{i}|z_{t-1}^{i},y_{t}^{i})\right|o=1\right]\nonumber\\
	&=\frac{1}{p(o=1;\bs{\theta}_{k})}\sum_{T=1}^{\infty}\sum_{x_{0:T},\bs{y}_{0:T},\bs{z}_{0:T},\bs{a}_{0:T}}p(o=1,x_{0:T},\bs{y}_{0:T},\bs{z}_{0:T},\bs{a}_{0:T},T;\bs{\theta}_{k})\sum_{t=1}^{T}\log \lambda^{i}(z_{t}^{i}|z_{t-1}^{i},y_{t}^{i})\nonumber\\
	&=\frac{1}{p(o=1;\bs{\theta}_{k})}\sum_{T=1}^{\infty}\sum_{t=1}^{T}\sum_{x_{t-1:t},\bs{y}_{t},\bs{z}_{t-1:t}}p(o=1,x_{t-1:t},\bs{y}_{t},\bs{z}_{t-1:t},T;\bs{\theta}_{k})\log \lambda^{i}(z_{t}^{i}|z_{t-1}^{i},y_{t}^{i})\nonumber\\
	&=\frac{1}{p(o=1;\bs{\theta}_{k})}\sum_{T=1}^{\infty}\sum_{t=1}^{T}\sum_{x_{t-1:t},\bs{y}_{t},\bs{z}_{t-1:t}}p(T)p_{t-1}(x_{t-1},\bs{z}_{t-1};\bs{\theta}_{k})\nonumber\\
	&\times p(x_{t},\bs{y}_{t}|x_{t-1},\bs{z}_{t-1};\bs{\pi}_{k})\bs{\lambda}_{k}(\bs{z}_{t}|\bs{z}_{t-1},\bs{y}_{t})p_{t}(o=1|x_{t},\bs{z}_{t},T;\bs{\theta}_{k})\log \lambda^{i}(z_{t}^{i}|z_{t-1}^{i},y_{t}^{i})\nonumber\\
	&=\frac{1}{p(o=1;\bs{\theta}_{k})}\sum_{x,x',\bs{y}',\bs{z},\bs{z}'}\bs{\lambda}_{k}(\bs{z}'|\bs{z},\bs{y}')\log \lambda^{i}(z^{i'}|z^{i},y^{i'})\nonumber\\
	&\times p(x',\bs{y}'|x,\bs{z};\bs{\pi}_{k})\sum_{T=1}^{\infty}p(T)\sum_{t=1}^{T}p_{t-1}(x,\bs{z};\bs{\theta}_{k})
	p_{t}(o=1|x',\bs{z}',T;\bs{\theta}_{k}).
\end{align}
Since $\sum_{T=1}^{\infty}\sum_{t=1}^{T}...=\sum_{t=1}^{\infty}\sum_{T=t}^{\infty}...$, we have
\begin{align}
	Q(\lambda^{i};\bs{\theta}_{k})
	=&\frac{1}{p(o=1;\bs{\theta}_{k})}\sum_{x,x',\bs{y}',\bs{z},\bs{z}'}\bs{\lambda}_{k}(\bs{z}'|\bs{z},\bs{y}')\log \lambda^{i}(z^{i'}|z^{i},y^{i'})\nonumber\\
	&\times p(x',\bs{y}'|x,\bs{z};\bs{\pi}_{k})\sum_{t=1}^{\infty}p_{t-1}(x,\bs{z};\bs{\theta}_{k})\sum_{T=t}^{\infty}p(T)p_{t}(o=1|x',\bs{z}',T;\bs{\theta}_{k})\nonumber\\
	=&\frac{1-\gamma}{p(o=1;\bs{\theta}_{k})}\sum_{x,x',\bs{y}',\bs{z},\bs{z}'}\bs{\lambda}_{k}(\bs{z}'|\bs{z},\bs{y}')\log \lambda^{i}(z^{i'}|z^{i},y^{i'})\nonumber\\
	&\times p(x',\bs{y}'|x,\bs{z};\bs{\pi}_{k})\sum_{t=1}^{\infty}p_{t-1}(x,\bs{z};\bs{\theta}_{k})\sum_{T=t}^{\infty}\gamma^{T}p_{t}(o=1|x',\bs{z}',T;\bs{\theta}_{k}).
\end{align}
Since $p_{t}(o=1,x,\bs{z},T;\bs{\theta}_{k})=p_{0}(o=1|x,\bs{z},T-t;$ $\bs{\theta}_{k})$, 
\begin{align}
	Q(\lambda^{i};\bs{\theta}_{k})
	=&\frac{1-\gamma}{p(o=1;\bs{\theta}_{k})}\sum_{x,x',\bs{y}',\bs{z},\bs{z}'}\bs{\lambda}_{k}(\bs{z}'|\bs{z},\bs{y}')\log \lambda^{i}(z^{i'}|z^{i},y^{i'})p(x',\bs{y}'|x,\bs{z};\bs{\pi}_{k})\nonumber\\
	&\times\sum_{t=1}^{\infty}p_{t-1}(x,\bs{z};\bs{\theta}_{k})\sum_{T=t}^{\infty}\gamma^{T}p_{0}(o=1|x',\bs{z}',T-t;\bs{\theta}_{k})\nonumber\\
	=&\frac{(1-\gamma)\gamma}{p(o=1;\bs{\theta}_{k})}\sum_{x,x',\bs{y}',\bs{z},\bs{z}'}\bs{\lambda}_{k}(\bs{z}'|\bs{z},\bs{y}')\log \lambda^{i}(z^{i'}|z^{i},y^{i'})p(x',\bs{y}'|x,\bs{z};\bs{\pi}_{k})\nonumber\\
	&\times\sum_{t=0}^{\infty}\gamma^{t}p_{t}(x,\bs{z};\bs{\theta}_{k})\sum_{T=0}^{\infty}\gamma^{T}p_{0}(o=1|x',\bs{z}',T;\bs{\theta}_{k})\nonumber\\
	=&\frac{(1-\gamma)\gamma}{p(o=1;\bs{\theta}_{k})}\sum_{x,x',\bs{y}',\bs{z},\bs{z}'}\bs{\lambda}_{k}(\bs{z}'|\bs{z},\bs{y}')\log \lambda^{i}(z^{i'}|z^{i},y^{i'})\nonumber\\
	&\times p(x',\bs{y}'|x,\bs{z};\bs{\pi}_{k})
	F(x,\bs{z};\bs{\theta}_{k})V(x',\bs{z}';\bs{\theta}_{k}).
\end{align}
Therefore, Equation (\ref{eq: Q(lambda)}) is proved. \\

Finally, we prove Equation (\ref{eq: Q(nu)}). $Q(\nu^{i};\bs{\theta}_{k})$ can be calculated as follows: 
\begin{align}
	&Q(\nu^{i};\bs{\theta}_{k})
	:=\mb{E}_{\bs{\theta}_{k}}\left[\left.\log \nu^{i}(z_{0}^{i})\right|o=1\right]\nonumber\\
	&=\frac{1}{p(o=1;\bs{\theta}_{k})}\sum_{T=0}^{\infty}\sum_{x_{0:T},\bs{y}_{0:T},\bs{z}_{0:T},\bs{a}_{0:T}}p(o=1,x_{0:T},\bs{y}_{0:T},\bs{z}_{0:T},\bs{a}_{0:T},T;\bs{\theta}_{k})\log \nu^{i}(z_{0}^{i})\nonumber\\
	&=\frac{1}{p(o=1;\bs{\theta}_{k})}\sum_{T=0}^{\infty}\sum_{x_{0},\bs{z}_{0}}p(o=1,x_{0},\bs{z}_{0},T;\bs{\theta}_{k})\log \nu^{i}(z_{0}^{i})\nonumber\\
	&=\frac{1}{p(o=1;\bs{\theta}_{k})}\sum_{T=0}^{\infty}\sum_{x_{0},\bs{z}_{0}}p(T)p_{0}(x_{0})\bs{\nu}_{k}(\bs{z}_{0})p_{0}(o=1|x_{0},\bs{z}_{0},T;\bs{\theta}_{k})\log \nu^{i}(z_{0}^{i})\nonumber\\
	&=\frac{1}{p(o=1;\bs{\theta}_{k})}\sum_{T=0}^{\infty}\sum_{x,\bs{z}}p(T)p_{0}(x)\bs{\nu}_{k}(\bs{z})p_{0}(o=1|x,\bs{z},T;\bs{\theta}_{k})\log \nu^{i}(z^{i})\nonumber\\
	&=\frac{1}{p(o=1;\bs{\theta}_{k})}\sum_{\bs{z}}\bs{\nu}_{k}(\bs{z})\log \nu^{i}(z^{i})\sum_{x}p_{0}(x)\sum_{T=0}^{\infty}p(T)p_{0}(o=1|x,\bs{z},T;\bs{\theta}_{k})\nonumber\\
	&=\frac{1-\gamma}{p(o=1;\bs{\theta}_{k})}\sum_{\bs{z}}\bs{\nu}_{k}(\bs{z})\log \nu^{i}(z^{i})\sum_{x}p_{0}(x)V(x,\bs{z};\bs{\theta}_{k}).
\end{align}
Therefore, Equation (\ref{eq: Q(nu)}) is proved. 
\end{proof}

Equations (\ref{Eq: M step-pi})--(\ref{Eq: M step-nu}) can be calculated from Equations (\ref{eq: Q(pi)})--(\ref{eq: Q(nu)}) using the Lagrange multiplier method \cite{kumar2010}. 
Therefore, Proposition \ref{th: M step} is proved. 

\subsection{Proof of Proposition \ref{prop: Tmax}}\label{adx: Tmax}
The left-hand side of Equation (\ref{eq: error of F in EM}) can be calculated as follows: 
\begin{align}
	&\left\|F(x,\bs{z};\bs{\theta}_{k})-\sum_{t=0}^{T_{\max}}\gamma^{t}\alpha_{t}(x,\bs{z};\bs{\theta}_{k})\right\|_{\infty}\nonumber\\
	=&\left\|\sum_{t=T_{\max}+1}^{\infty}\gamma^{t}\alpha_{t}(x,\bs{z};\bs{\theta}_{k})\right\|_{\infty}
	\leq\sum_{t=T_{\max}+1}^{\infty}\gamma^{t}\left\|\alpha_{t}(x,\bs{z};\bs{\theta}_{k})\right\|_{\infty}
	=\sum_{t=T_{\max}+1}^{\infty}\gamma^{t}
	=\frac{\gamma^{T_{\max}+1}}{1-\gamma}.
\end{align}
Therefore, if 
\begin{align}
    \frac{\gamma^{T_{\max}+1}}{1-\gamma}<\varepsilon
    \label{eq: mistake in EM tmp}
\end{align}
is satisfied, Equation (\ref{eq: error of F in EM}) is satisfied. From Equation (\ref{eq: mistake in EM tmp}), we obtain Equation (\ref{eq: Tmax}). Therefore, Equation (\ref{eq: Tmax})$\Rightarrow$Equation (\ref{eq: error of F in EM}) is proved. 
Equation (\ref{eq: Tmax})$\Rightarrow$Equation (\ref{eq: error of V in EM}) can be proved in the same way. 

\section{Proof in Section~\ref{section: Bellman EM algorithm}}
\subsection{Proof of Theorem \ref{th: bellman equations}}\label{adx: bellman equations}
$F(x,\bs{z};\bs{\theta})$ can be calculated as follows: 
\begin{align}
	F(x,\bs{z};\bs{\theta})
	:=&\sum_{t=0}^{\infty}\gamma^{t}p_{t}(x,\bs{z};\bs{\theta})\nonumber\\
	=&p_{0}(x,\bs{z};\bs{\nu})+\sum_{t=1}^{\infty}\gamma^{t}p_{t}(x,\bs{z};\bs{\theta})\nonumber\\
	=&p_{0}(x,\bs{z};\bs{\nu})+\sum_{t=1}^{\infty}\gamma^{t}\sum_{x',\bs{z}'}p(x,\bs{z}|x',\bs{z}';\bs{\theta})p_{t-1}(x',\bs{z}';\bs{\theta})\nonumber\\
	=&p_{0}(x,\bs{z};\bs{\nu})+\gamma\sum_{x',\bs{z}'}p(x,\bs{z}|x',\bs{z}';\bs{\theta})\sum_{t=1}^{\infty}\gamma^{t-1}p_{t-1}(x',\bs{z}';\bs{\theta})\nonumber\\
	=&p_{0}(x,\bs{z};\bs{\nu})+\gamma\sum_{x',\bs{z}'}p(x,\bs{z}|x',\bs{z}';\bs{\theta})F(x',\bs{z}';\bs{\theta}).
\end{align}
Therefore, Equation (\ref{eq: FBE}) is proved. 
Equation (\ref{eq: BBE}) can be proved in the same way. 

\section{Proof in Section~\ref{section: Modified Bellman EM algorithm}}
\subsection{Proof of Proposition \ref{prop: contractibility of Bellman operators}}\label{adx: contractibility of Bellman operators}
The left-hand side of Equation (\ref{eq: contractibility of A}) can be calculated as follows: 
\begin{align}
    \|A_{\bs{\theta}}f-A_{\bs{\theta}}g\|_{1}
    &=\gamma\|\bs{P}(\bs{\theta})\bs{f}-\bs{P}(\bs{\theta})\bs{g}\|_{1}\nonumber\\
    &\leq\gamma\|\bs{P}(\bs{\theta})\|_{1}\|\bs{f}-\bs{g}\|_{1}\nonumber\\
    &=\gamma\|f-g\|_{1}.
\end{align}
where $P_{ij}(\bs{\theta}):=p((x',\bs{z}')=i|(x,\bs{z})=j;\bs{\theta})$, $f_{i}:=f((x,\bs{z})=i)$, and $g_{i}:=g((x,\bs{z})=i)$. Equation (\ref{eq: contractibility of B}) can be proved in the same way. 

\subsection{Proof of Proposition \ref{prop: F and V from Bellman operators}}\label{adx: F and V from Bellman operators}
We prove Equation (\ref{eq: FV derived from BO}) by showing $\lim_{L\to\infty}\|F(\bs{\theta})-A_{\bs{\theta}}^{L}f\|_{1}=0$. 
Since $F(x,\bs{z};\bs{\theta})$ is the fixed point of $A_{\bs{\theta}}$, 
\begin{align}
    \|F(\bs{\theta})-A_{\bs{\theta}}^{L}f\|_{1}&=\|A_{\bs{\theta}}^{L}F-A_{\bs{\theta}}^{L}f\|_{1}.
\end{align}
From the contractibility of $A_{\bs{\theta}}$, 
\begin{align}
    \|A_{\bs{\theta}}^{L}F(\bs{\theta})-A_{\bs{\theta}}^{L}f\|_{1}\leq\gamma^{L}\|F(\bs{\theta})-f\|_{1}.
\end{align}
Since $\|F(\bs{\theta})-f\|_{1}$ is finite, the right-hand side is $0$ when $L\to\infty$. Hence, 
\begin{align}
    \lim_{L\to\infty}\|F(\bs{\theta})-A_{\bs{\theta}}^{L}f\|_{1}&=0
\end{align}
is satisfied, and Equation (\ref{eq: FV derived from BO}) is proved. 
Equation (\ref{eq: BV derived from BO}) can be proved in the same way. 

\subsection{Proof of Proposition \ref{prop: the termination condition by the contractibility of the Bellman operators}}\label{adx: the termination condition by the contractibility of the Bellman operators}
From the definition of the norm, 
\begin{align}
    \left\|F(\bs{\theta}_{k})-A_{\bs{\theta}_{k}}^{L}f\right\|_{\infty}&\leq\left\|F(\bs{\theta}_{k})-A_{\bs{\theta}_{k}}^{L}f\right\|_{1}.
\end{align}
The right-hand side can be calculated as follows:
\begin{align}
	\left\|F(\bs{\theta}_{k})-A_{\bs{\theta}_{k}}^{L}f\right\|_{1}
	\leq&\left\|F(\bs{\theta}_{k})-A_{\bs{\theta}_{k}}^{L+1}f\right\|_{1}+\left\|A_{\bs{\theta}_{k}}^{L+1}f-A_{\bs{\theta}_{k}}^{L}f\right\|_{1}\nonumber\\
	=&\left\|A_{\bs{\theta}_{k}}F(\bs{\theta}_{k})-A_{\bs{\theta}_{k}}^{L+1}f\right\|_{1}+\left\|A_{\bs{\theta}_{k}}^{L+1}f-A_{\bs{\theta}_{k}}^{L}f\right\|_{1}\nonumber\\
	\leq&\gamma\left\|F(\bs{\theta}_{k})-A_{\bs{\theta}_{k}}^{L}f\right\|_{1}+\gamma\left\|A_{\bs{\theta}_{k}}^{L}f-A_{\bs{\theta}_{k}}^{L-1}f\right\|_{1}.
\end{align}
From this inequality, we have
\begin{align}
    \left\|F(\bs{\theta}_{k})-A_{\bs{\theta}_{k}}^{L}f\right\|_{1}
    \leq&\frac{\gamma}{1-\gamma}\left\|A_{\bs{\theta}_{k}}^{L}f-A_{\bs{\theta}_{k}}^{L-1}f\right\|_{1}.
\end{align}
Therefore, if $\|A_{\bs{\theta}_{k}}^{L}f-A_{\bs{\theta}_{k}}^{L-1}f\|_{1}<(1-\gamma)\gamma^{-1}\varepsilon$ is satisfied, 
\begin{align}
    \left\|F(\bs{\theta}_{k})-A_{\bs{\theta}_{k}}^{L}f\right\|_{\infty}&<\varepsilon
\end{align}
holds. Equation (\ref{eq: approx error of MBEM-B}) is proved in the same way. 

\subsection{Proof of Proposition \ref{prop: L is smaller than T by the contractibility of the Bellman operators}}\label{adx: L is smaller than T by the contractibility of the Bellman operators}
When $f(x,\bs{z})=p_{0}(x,\bs{z};\bs{\nu}_{k})$ and $v(x,\bs{z})=\bar{r}(x,\bs{z};\bs{\pi}_{k})$, 
the termination conditions of MBEM can be calculated as follows:
\begin{align}
	\left\|A_{\bs{\theta}_{k}}^{L}f-A_{\bs{\theta}_{k}}^{L-1}f\right\|_{1}
	&\leq\gamma^{L-1}\left\|A_{\bs{\theta}_{k}}f-f\right\|_{1}\nonumber\\
	&=\gamma^{L-1}\left\|\gamma\bs{P}(\bs{\theta}_{k})\bs{p}(\bs{\nu}_{k})\right\|_{1}
	\leq\gamma^{L}\left\|\bs{P}(\bs{\theta}_{k})\right\|_{1}\left\|\bs{p}(\bs{\nu}_{k})\right\|_{1}
	=\gamma^{L},\\
	\left\|B_{\bs{\theta}_{k}}^{L}v-B_{\bs{\theta}_{k}}^{L-1}v\right\|_{\infty}
	&\leq\gamma^{L}.
\end{align}
The calculation of $\|B_{\bs{\theta}_{k}}^{L}v-B_{\bs{\theta}_{k}}^{L-1}v\|_{\infty}$ is omitted because it is the same as that of $\|A_{\bs{\theta}_{k}}^{L}f-A_{\bs{\theta}_{k}}^{L-1}f\|_{1}$. 
If 
\begin{align}
	\gamma^{L}<\frac{1-\gamma}{\gamma}\varepsilon
	\ \ \ \left(\Rightarrow
	L>\frac{\log(1-\gamma)\varepsilon}{\log\gamma}-1\right)
	\label{eq: 1568}
\end{align}
is satisfied, 
\begin{align}
	&\|A_{\bs{\theta}_{k}}^{L}f-A_{\bs{\theta}_{k}}^{L-1}f\|_{1}<(1-\gamma)\gamma^{-1}\varepsilon,\\
	&\|B_{\bs{\theta}_{k}}^{L}v-B_{\bs{\theta}_{k}}^{L-1}v\|_{\infty}<(1-\gamma)\gamma^{-1}\varepsilon
\end{align}
holds. Thus, $L_{\max}$ satisfies
\begin{align}
	L_{\max}\leq\left\lceil\frac{\log(1-\gamma)\varepsilon}{\log\gamma}-1\right\rceil.
\end{align}
$\lceil\cdot\rceil: \mb{R}\to\mb{Z}$ is defined by $\lceil x\rceil:=\min\{n\in\mb{Z}|n>x\}$. 
From Proposition \ref{prop: Tmax}, the minimum $T_{\max}$ is given by the following equation: 
\begin{align}
	T_{\max}=\left\lceil\frac{\log(1-\gamma)\varepsilon}{\log\gamma}-1\right\rceil.
\end{align}
Therefore, $L_{\max}\leq T_{\max}$ holds.

\section{A Note on the Algorithm proposed by Song et al.}\label{adx: mistake in Song2016}
In this section, we show that a parameter dependency is overlooked in the algorithm of \cite{Song2016}. 
We outline the derivation of the algorithm in \cite{Song2016} and discuss the parameter dependency. 
Since we use the notation in this paper, it is recommended to read the full paper before reading this section. 

Firstly, we calculate the expected return $J(\bs{\theta})$ to derive the algorithm in \cite{Song2016}. 
Since \cite{Song2016} considers the case where $\bs{\nu}(\bs{z})=\delta_{\bs{z},\bs{z}_{0}}$, 
we also consider the same case in this section.
The expected return $J(\bs{\theta})$ can be calculated as follows:
\begin{align}
	J(\bs{\theta})
	:=&\mb{E}_{\bs{\theta}}\left[\sum_{t=0}^{\infty}\gamma^{t}r(x_{t},\bs{a}_{t})\right]\nonumber\\
	=&\sum_{x_{0},\bs{a}_{0}}p(x_{0})\bs{\pi}(\bs{a}_{0}|\bs{z}_{0})\nonumber\\
	\times&\left[r(x_{0},\bs{a}_{0})+\gamma\sum_{x_{1},\bs{y}_{1},\bs{z}_{1}}p(x_{1}|x_{0},\bs{a}_{0})p(\bs{y}_{1}|x_{1},\bs{a}_{0})\bs{\lambda}(\bs{z}_{1}|\bs{z}_{0},\bs{y}_{1})V(x_{1},\bs{z}_{1};\bs{\theta})\right].
	\label{eq: expected return 1290}
\end{align}
$V(x,\bs{z};\bs{\theta})$ is the value function, which is defined as follows: 
\begin{align}
	V(x,\bs{z};\bs{\theta}):=\mb{E}_{\bs{\theta}}\left[\left.\sum_{t=0}^{\infty}\gamma^{t}r(x_{t},\bs{a}_{t})\right|\bs{x}_{0}=x,\bs{z}_{0}=\bs{z}\right].
\end{align}
Equation (\ref{eq: expected return 1290}) can be rewritten as: 
\begin{align}
	J(\bs{\theta})
	=&\sum_{a_{0}^{i}}\pi^{i}(a_{0}^{i}|z_{0}^{i})\tilde{r}(z_{0}^{i},a_{0}^{i};\bs{\pi}^{-i})
	+\sum_{a_{0}^{i},y_{1}^{i},z_{1}^{i}}\pi^{i}(a_{0}^{i}|z_{0}^{i})\lambda^{i}(z_{1}^{i}|z_{0}^{i},y_{1}^{i})\tilde{V}(z_{0}^{i},a_{0}^{i},y_{1}^{i},z_{1}^{i};\bs{\theta})
\end{align}
where
\begin{align}
	\tilde{r}(z_{0}^{i},a_{0}^{i};\bs{\pi}^{-i})
	:=&\sum_{x_{0},\bs{a}_{0}^{-i}}p(x_{0})\bs{\pi}^{-i}(\bs{a}_{0}^{-i}|\bs{z}_{0}^{-i})r(x_{0},\bs{a}_{0}),\\
	\tilde{V}(z_{0}^{i},a_{0}^{i},y_{1}^{i},z_{1}^{i};\bs{\theta})
	:=&\gamma\sum_{x_{0},\bs{a}_{0}^{-i},x_{1},\bs{y}_{1}^{-i},\bs{z}_{1}^{-i}}p(x_{0})\bs{\pi}^{-i}(\bs{a}_{0}^{-i}|\bs{z}_{0}^{-i})p(x_{1}|x_{0},\bs{a}_{0})\nonumber\\
	&\times p(\bs{y}_{1}|x_{1},\bs{a}_{0})\bs{\lambda}^{-i}(\bs{z}_{1}^{-i}|\bs{z}_{0}^{-i},\bs{y}_{1}^{-i})V(x_{1},\bs{z}_{1};\bs{\theta}).
\end{align}
Maximizing $J(\bs{\theta})$ is equivalent to maximizing $\log J(\bs{\theta})$, and $\log J(\bs{\theta})$ can be calculated as follows: 
\begin{align}
	\log J(\bs{\theta})
	=&\log\left\{\sum_{a_{0}^{i}}\eta(z_{0}^{i},a_{0}^{i};\bs{\theta}_{k})\frac{\pi^{i}(a_{0}^{i}|z_{0}^{i})\tilde{r}(z_{0}^{i},a_{0}^{i};\bs{\pi}^{-i})}{\eta(z_{0}^{i},a_{0}^{i};\bs{\theta}_{k})}\right.\nonumber\\
	&\left.+\sum_{a_{0}^{i},y_{1}^{i},z_{1}^{i}}\rho(z_{0}^{i},a_{0}^{i},y_{1}^{i},z_{1}^{i};\bs{\theta}_{k})\frac{\pi^{i}(a_{0}^{i}|z_{0}^{i})\lambda^{i}(z_{1}^{i}|z_{0}^{i},y_{1}^{i})\tilde{V}(z_{0}^{i},a_{0}^{i},y_{1}^{i},z_{1}^{i};\bs{\theta})}{\rho(z_{0}^{i},a_{0}^{i},y_{1}^{i},z_{1}^{i};\bs{\theta}_{k})}\right\}
	\label{eq: log expected return 1319}
\end{align}
where
\begin{align}
	\eta(z_{0}^{i},a_{0}^{i};\bs{\theta}_{k})
	=&\frac{\pi^{i}_{k}(a_{0}^{i}|z_{0}^{i})\tilde{r}(z_{0}^{i},a_{0}^{i};\bs{\pi}^{-i}_{k})}{\xi(z_{0}^{i};\bs{\theta}_{k})},\label{eq: eta of E step in Song}\\
	\rho(z_{0}^{i},a_{0}^{i},y_{1}^{i},z_{1}^{i};\bs{\theta}_{k})=&\frac{\pi^{i}_{k}(a_{0}^{i}|z_{0}^{i})\lambda^{i}_{k}(z_{1}^{i}|z_{0}^{i},y_{1}^{i})\tilde{V}(z_{0}^{i},a_{0}^{i},y_{1}^{i},z_{1}^{i};\bs{\theta}_{k})}{\xi(z_{0}^{i};\bs{\theta}_{k})},\label{eq: rho of E step in Song}\\
	\xi(z_{0}^{i};\bs{\theta}_{k})
	=&\sum_{a_{0}^{i}}\pi^{i}_{k}(a_{0}^{i}|z_{0}^{i})\tilde{r}(z_{0}^{i},a_{0}^{i};\bs{\pi}^{-i}_{k})\nonumber\\
	&+\sum_{a_{0}^{i},y_{1}^{i},z_{1}^{i}}\pi^{i}_{k}(a_{0}^{i}|z_{0}^{i})\lambda^{i}_{k}(z_{1}^{i}|z_{0}^{i},y_{1}^{i})\tilde{V}(z_{0}^{i},a_{0}^{i},y_{1}^{i},z_{1}^{i};\bs{\theta}_{k}).\label{eq: xi of E step in Song}
\end{align}
By the Jensen's inequality, Equation (\ref{eq: log expected return 1319}) can be calculated as follows: 
\begin{align}
	\log J(\bs{\theta})
	\geq&\sum_{a_{0}^{i}}\eta(z_{0}^{i},a_{0}^{i};\bs{\theta}_{k})\log\left\{\frac{\pi^{i}(a_{0}^{i}|z_{0}^{i})\tilde{r}(z_{0}^{i},a_{0}^{i};\bs{\pi}^{-i})}{\eta(z_{0}^{i},a_{0}^{i};\bs{\theta}_{k})}\right\}\nonumber\\
	+&\sum_{a_{0}^{i},y_{1}^{i},z_{1}^{i}}\rho(z_{0}^{i},a_{0}^{i},y_{1}^{i},z_{1}^{i};\bs{\theta}_{k})\log\left\{\frac{\pi^{i}(a_{0}^{i}|z_{0}^{i})\lambda^{i}(z_{1}^{i}|z_{0}^{i},y_{1}^{i})\tilde{V}(z_{0}^{i},a_{0}^{i},y_{1}^{i},z_{1}^{i};\bs{\theta})}{\rho(z_{0}^{i},a_{0}^{i},y_{1}^{i},z_{1}^{i};\bs{\theta}_{k})}\right\}\nonumber\\
	=:&Q(\bs{\theta};\bs{\theta}_{k}).	
	\label{eq: inequality of log expected return 1337}
\end{align}
$\log J(\bs{\theta})=Q(\bs{\theta};\bs{\theta}_{k})$ is satisfied when $\bs{\theta}=\bs{\theta}_{k}$. 

Then, $\bs{\theta}_{k+1}$ is defined as follows: 
\begin{align}
	\bs{\theta}_{k+1}:=\arg\max_{\bs{\theta}}Q(\bs{\theta};\bs{\theta}_{k})\label{eq: M step in Song}
\end{align}
In this case, the following proposition holds: 

\begin{Proposition}[\cite{Song2016}]
$\log J(\bs{\theta}_{k+1})\geq \log J(\bs{\theta}_{k})$.
\end{Proposition}

\begin{proof}
From Equation (\ref{eq: inequality of log expected return 1337}), $\log J(\bs{\theta}_{k+1})\geq Q(\bs{\theta}_{k+1};\bs{\theta}_{k})$. 
From Equation (\ref{eq: M step in Song}), $Q(\bs{\theta}_{k+1};\bs{\theta}_{k})\geq Q(\bs{\theta}_{k};\bs{\theta}_{k})=\log J(\bs{\theta}_{k})$.
Therefore, $\log J(\bs{\theta}_{k+1})\geq \log J(\bs{\theta}_{k})$ is satisfied.
\end{proof}

Therefore, since Equation (\ref{eq: M step in Song}) monotonically increases $J(\bs{\theta})$, we can find $\bs{\theta}^{*}$, which locally maximizes $J(\bs{\theta})$. 
This is the algorithm in \cite{Song2016}. 

Then, the problem is how to calculate Equation (\ref{eq: M step in Song}). 
It cannot be calculated analytically 
because $\bs{\theta}$ dependency of $\tilde{V}(z_{0}^{i},a_{0}^{i},y_{1}^{i},z_{1}^{i};\bs{\theta})$ is too complex. 
However, \cite{Song2016} overlooked the parameter dependency of $\tilde{r}(z_{0}^{i},a_{0}^{i};\bs{\pi}^{-i})$ and $\tilde{V}(z_{0}^{i},a_{0}^{i},y_{1}^{i},z_{1}^{i};\bs{\theta})$, 
and therefore, it calculated Equation (\ref{eq: M step in Song}) as follows: 
\begin{align}
	&\pi_{k+1}^{i}(a_{0}^{i}|z_{0}^{i})=\frac{\eta(z_{0}^{i},a_{0}^{i};\bs{\theta}_{k})+\sum_{y_{1}^{i},z_{1}^{i}}\rho(z_{0}^{i},a_{0}^{i},y_{1}^{i},z_{1}^{i};\bs{\theta}_{k})}
	{\sum_{a_{0}^{i}}\left[\eta(z_{0}^{i},a_{0}^{i};\bs{\theta}_{k})+\sum_{y_{1}^{i},z_{1}^{i}}\rho(z_{0}^{i},a_{0}^{i},y_{1}^{i},z_{1}^{i};\bs{\theta}_{k})\right]},\label{eq: incorrect M step pi in Song}\\
	&\lambda_{k+1}^{i}(z_{1}^{i}|z_{0}^{i},y_{1}^{i})=\frac{\sum_{a_{0}^{i}}\rho(z_{0}^{i},a_{0}^{i},y_{1}^{i},z_{1}^{i};\bs{\theta}_{k})}
	{\sum_{a_{0}^{i},z_{1}^{i}}\rho(z_{0}^{i},a_{0}^{i},y_{1}^{i},z_{1}^{i};\bs{\theta}_{k})}.\label{eq: incorrect M step lambda in Song}
\end{align}
However, Equations (\ref{eq: incorrect M step pi in Song}) and (\ref{eq: incorrect M step lambda in Song}) do not correspond to Equation (\ref{eq: M step in Song}), 
and therefore, the algorithm as a whole may not always provide the optimal policy.

\end{paracol}
\reftitle{References}

\end{document}